%% file: main.tex
\documentclass{article}
\usepackage{iclr2027_conference,times}

\input{math_commands.tex}

\usepackage{url}
\usepackage{amsmath,amssymb,amsthm}
\usepackage{booktabs}
\usepackage{mathtools}
\usepackage{wrapfig}
\usepackage{graphicx}
\usepackage{array}
\usepackage{longtable}

\usepackage{tikz}
\usepackage{xcolor}
\usepackage{placeins}
\AtBeginDocument{\setlength{\LTcapwidth}{\linewidth}}
\newif\ifrevision \revisionfalse
\ifrevision
  \newcommand{\rev}[1]{{\color{red!80!black}#1}}
  
\else
  \newcommand{\rev}[1]{#1}
  
\fi
\usetikzlibrary{positioning,arrows.meta,calc,fit,backgrounds,shapes.geometric}

\newcommand{\Unif}{\operatorname{Unif}}

\title{Test-Time Symmetrization: Scaling Laws for Model-Agnostic Learning under Structure}

\author{Behrooz Tahmasebi \& Melanie Weber \\
Harvard John A.~Paulson School of Engineering and Applied Sciences (SEAS)\\
Harvard  University, Cambridge, MA 02138, USA \\
\texttt{\{behrooz\_tahmasebi,mweber\}@seas.harvard.edu} 
}

\definecolor{darkblue}{rgb}{0.0,0.0,0.65}
\definecolor{darkred}{rgb}{0.68,0.05,0.0}
\definecolor{darkgreen}{rgb}{0.0,0.29,0.29}
\definecolor{darkpurple}{rgb}{0.47,0.09,0.29}

\usepackage[backref=page]{hyperref}
\hypersetup{
  colorlinks=true,
  citecolor=darkblue,
  linkcolor=darkred,
  filecolor=darkblue,
  urlcolor=darkblue
}

\usepackage[capitalize,noabbrev]{cleveref}
\crefname{corollary}{Corollary}{Corollaries}
\Crefname{corollary}{Corollary}{Corollaries}

\newtheorem{theorem}{Theorem}

\newtheorem{corollary}[theorem]{Corollary}
\newtheorem{lemma}[theorem]{Lemma}
\theoremstyle{definition}
\newtheorem{definition}[theorem]{Definition}
\theoremstyle{remark}

\begin{document}

\maketitle
\addtocontents{toc}{\protect\setcounter{tocdepth}{-10}}

\begin{abstract} 
    Test-time compute has emerged as a powerful way to improve the accuracy of language models at inference time, with scaling laws quantifying how accuracy improves with additional forward passes. Existing scaling laws, however, focus on repeated sampling from a fixed prompt, even though many tasks admit multiple equivalent representations of the same underlying problem.
    A graph does not depend on how its nodes are numbered, and a program does not depend on the names of its variables, yet language models can change their answers under such rewritings. 
    In this paper, we study \emph{test-time symmetrization}, which uses forward passes on rewritings of the input and aggregates the resulting answers over the group $G$ of rewritings.
    We prove that no amount of sampling and repeated decoding from a single representation can replace rewriting, while only $O(\log |G|)$ random rewritings suffice to recover the information provided by all $|G|$ rewritings.
    The same set of $O(\log |G|)$ rewritings provides simultaneous approximation across models, decoding rules, and sampled reasoning traces, and the logarithmic scaling is optimal in general. \rev{We validate the theory with experiments on Qwen2.5-7B, Phi-4-mini, Llama-3.1-8B, and Qwen2.5-14B, with chain-of-thought reasoning, on graph reasoning, logical deduction, and multiple-choice question answering. At the same number of forward passes, spending them on rewritings of the input is substantially more accurate than spending them on repeated sampling from the original prompt whenever the model's answer depends on how the input is written, and the number of rewritings needed grows only logarithmically with the number of possible rewritings, as the theory predicts.} Our results provide a principled way to enforce symmetries at inference time when they cannot be built into the model, which we call model-agnostic symmetrization.
\end{abstract}

\section{Introduction}
\label{sec:intro}

Test-time compute is now a standard way to improve a language model whose weights are fixed. A longer reasoning trace \citep{wei2022chain}, a majority vote over several traces \citep{wang2023selfconsistency}, or a verifier that selects among candidates \citep{cobbe2021verifiers,lightman2024verify} each convert extra forward passes into accuracy, and scaling laws predict the return on each pass \citep{brown2024monkeys,snell2025scaling}. All of these methods query the model again on the same prompt: the randomness they exploit is that of \emph{decoding}.

\begin{figure}[t]
    \centering
    \resizebox{\linewidth}{!}{%
        \input{figures/two-sources}
    }
    \caption{Rewriting by \(G\) explores the orbit of equivalent representations, while the frozen model \(p_\theta\) samples reasoning traces at each representation. Full orbit averaging uses \(|G|\) rewrites; test-time symmetrization uses only \(O(\log |G|)\) random rewrites with a few traces each.}
    \label{fig:two-sources}
\end{figure}
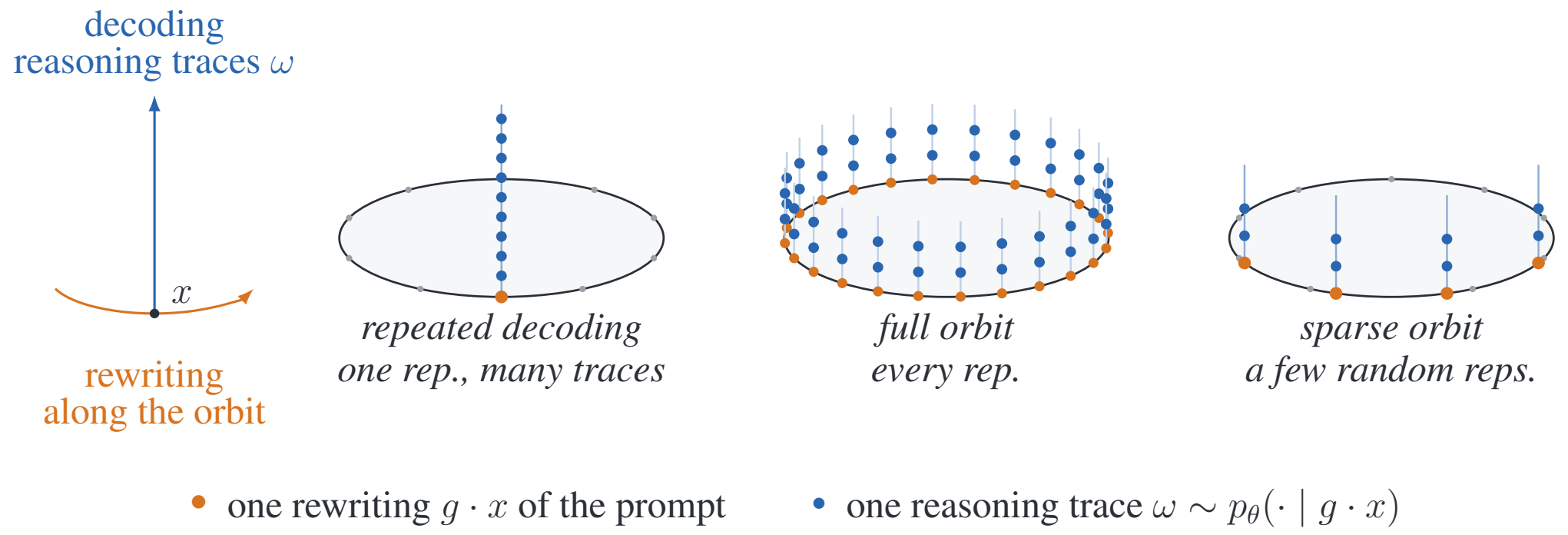

A second source of variation is \emph{rewriting}. Relabeling the nodes of a graph, permuting the options of a multiple-choice question, renaming the variables of a program, or reordering in-context examples changes the prompt but not its meaning. These rewritings move the input along an \emph{orbit} of equivalent representations, and a forward pass spent on a rewriting asks the same question in a different form. The two sources are not interchangeable. Repeated decoding explores the model at one representation and reveals nothing about its behavior elsewhere on the orbit (\cref{fig:two-sources}).

Indeed, frozen models are far from constant along the orbit. Reordering answer options or few-shot examples changes benchmark scores and model rankings \citep{zheng2024pride,lu2022fantastically,sclar2024quantifying,mizrahi2024state}, and the same instability appears for graph serializations and identifier names \citep{herbst2025lost,micelibarone2023larger}. Querying several rewritings and combining the answers helps: permutation-based calibration reduces option-order bias \citep{zheng2024pride}, prompt ensembles improve on single prompts \citep{jiang2020how,arora2023ama}, and at matched compute, paraphrasing the input can beat sampling more traces \citep{kozodoi2026augmentation}. These remedies are empirical, and three basic questions remain open: can more decoding replace rewriting, how many rewritings are enough, and how should rewriting be combined with decoding? \citet{kozodoi2026augmentation} point to the last one, the balance between input-side and output-side diversity, as a direction for future work. This paper takes up these questions.

\emph{Rewriting is necessary.} We construct two models whose responses agree on every trace at the original prompt but lead to different decisions once averaged over the orbit. No decoding procedure, adaptive or not, can tell them apart, however many traces it draws (\cref{thm:two-worlds}).

\emph{Logarithmically many rewritings suffice.} Note that a full pass over the group is out of reach: a graph on $20$ nodes has $20!\approx 2\times 10^{18}$ relabelings. We prove that a random set of rewritings whose size grows with $\log|G|$ approximates the full group average in mean square over starting representations, and that the logarithm cannot be improved in general (\cref{thm:sparse-reynolds}). The set is drawn once and serves every frozen model at the same time. This uniformity is what makes the guarantee \emph{model-agnostic}. Indeed, even though equivariant architectures build symmetry into the network during training, a frozen model that allows no such modification can still exploit task symmetries imposed externally through pre- and post-processing, at a logarithmic cost.

\emph{Any decoding rule can be used.} The same set of rewritings also serves every decoding rule. The traces at one rewriting may be combined by majority vote, expected logits, verifier weighting, or any other rule, at any generation temperature, and the result is again a function on the group. The scaling law holds for all such functions at once, including every realized randomized decoder, so the decoding rule can be chosen after the rewritings are drawn (\cref{thm:decoder-agnostic}). An error decomposition separates the accuracy of the fully symmetrized model from the cost of approximating it (\cref{cor:margin}), and the number of rewritings and the number of traces per rewriting can be set independently (\cref{app:joint-sampling}).

In short, in this paper, we make the following contributions:
\begin{itemize}
    \item We formulate test-time symmetrization, in which decoding and rewriting are two separate sources of test-time randomness, and prove that decoding alone cannot replace rewriting (\cref{thm:two-worlds}).
    \item We prove a tight scaling law: $O(\log|G|)$ random rewritings approximate the full group average, uniformly over models, decoding rules, and realized decoders (\cref{thm:sparse-reynolds,thm:decoder-agnostic}).
    \item \rev{We validate the theory with experiments on Qwen2.5-7B, Phi-4-mini, Llama-3.1-8B, and Qwen2.5-14B, with chain-of-thought reasoning, on graph reasoning, logical deduction, and multiple-choice question answering. At the same number of forward passes, rewriting the input is substantially more accurate than repeated sampling from the original prompt whenever the model's answer depends on how the input is written, exactly as the theory predicts (\cref{sec:experiments}).}
\end{itemize}

\section{Related Work}
\label{sec:related}

\paragraph{Test-time compute.}
Spending additional forward passes at inference is a standard way to improve a frozen language model: lengthening a single reasoning trace \citep{wei2022chain,muennighoff2025s1}, sampling several traces and voting \citep{wang2023selfconsistency}, reranking them with a verifier \citep{cobbe2021verifiers,lightman2024verify}, or searching over intermediate steps \citep{yao2023tree,snell2025scaling}. Empirical scaling laws for these procedures relate coverage or accuracy to the number of samples \citep{brown2024monkeys,wu2025inference,chen2024more,li2024agents}, and several works explain their functional form \citep{schaeffer2025monkeys,levi2025simple,liu2025rethinking} or their limitations under imperfect verification \citep{stroebl2026limits}; see \citet{welleck2024decoding} for a survey and \citet{kaplan2020scaling,hoffmann2022training} for the training-time counterparts. These procedures query the model using a fixed representation of the prompt, leaving the role of rewritings comparatively less studied. Our results show that this is a limitation in principle and that rewriting obeys its own scaling law, distinct from that of the decoding stage.

\paragraph{Sensitivity to rewritings.}
It is well documented that language models are not invariant to semantically neutral rewritings: the order of in-context examples \citep{lu2022fantastically,zhao2021calibrate}, prompt formatting \citep{sclar2024quantifying,mizrahi2024state}, the order of multiple-choice options \citep{zheng2024pride,pezeshkpour2024sensitivity,zong2024fool,gupta2024changing}, the order of premises \citep{chen2024premise}, the names and numbers in a math problem \citep{mirzadeh2025gsmsymbolic}, the position of information in a long context \citep{liu2024lost}, the serialization of a graph \citep{wang2023nlgraph,fatemi2024talk,herbst2025lost}, and identifier renaming in code \citep{micelibarone2023larger,wang2023recode} can all change the model's prediction; consistency across paraphrases has also been studied on its own \citep{elazar2021measuring}. Existing remedies use small collections of rewritings to estimate and remove option-position bias \citep{zheng2024pride}, aggregate predictions from several prompts \citep{jiang2020how,schick2021pet,arora2023ama}, modify input representations to make outputs order-independent \citep{mcilroyyoung2024order}, or train for robustness to permutations \citep{chen2025pearl}. Closest to our setting, \citet{kozodoi2026augmentation} show empirically that allocating forward passes to input paraphrases can outperform self-consistency at matched compute, motivating the allocation problem between input-side and output-side diversity. Our framework separates this decoding choice from the cost of enforcing symmetry. Related work on model-agnostic symmetrization and on the theory of invariance is discussed in \cref{app:related}.

\section{Test-Time Symmetrization}
\label{sec:formulation}

\paragraph{Rewriting orbits and decoding fibers.}
Let $\mathcal X$ denote the space of prompts and let $\mathcal C$ denote the set of answer classes, with $C=|\mathcal C|$. The target function $y^\star:\mathcal X\to\mathcal C$ assigns each prompt its correct answer. Let a finite group $G$ act on $\mathcal X$.\footnote{Most rewritings of interest are exact group actions once they are applied to the structured object behind the prompt (node labels, answer options, in-context examples, or identifiers) and the result is serialized in a fixed way, as in \cref{app:exp-setup}. \Cref{thm:sparse-reynolds,thm:decoder-agnostic} use only functions on $G$, so they apply to any family of rewritings indexed by $G$.} For a fixed semantic instance $x\in\mathcal X$, its \emph{orbit}
$G\cdot x=\{g\cdot x:g\in G\}$ contains all rewritings generated by the group action. A frozen model queried on $g\cdot x$ defines a conditional distribution $p_{\theta,\tau}(\cdot\mid g\cdot x)$ over reasoning traces. We write
\begin{equation}
    \omega\sim p_{\theta,\tau}(\cdot\mid g\cdot x),
    \qquad V_x(g,\omega)\in\mathcal Z,
    \label{eq:score-field}
\end{equation}
where $\tau$ is the token generation temperature, $\omega$ is a sampled trace, and $V_x(g,\omega)$ is its response representation in a measurable space $\mathcal Z$. It may contain an answer, logits, a verifier score, or the entire trace. Without loss of generality, we assume that the task is invariant, $y^\star(g\cdot x)=y^\star(x)$. If $G$ acts equivariantly on the answer coordinates, we first align them using the inverse output action; the same results then apply. Task invariance does not require $V_x(g,\omega)$ to be invariant. Its variation across $g$ is precisely what test time symmetrization aims to remove.

We use \emph{rewriting} and \emph{serialization} for the same representations, with a small distinction in emphasis: $g$ denotes the rewriting operation, while $g\cdot x$ is the resulting serialization of the prompt. The response field has two distinct coordinates. The group element $g$ selects a representation of the prompt, while $\omega$ selects a decoder outcome from $p_{\theta,\tau}(\cdot\mid g\cdot x)$. We call the conditional probability space
\[
    \bigl(\Omega,p_{\theta,\tau}(\cdot\mid g\cdot x)\bigr)
\]
the \emph{decoding fiber} at $g\cdot x$. Thus, an orbit contains the equivalent representations of one semantic instance, and each representation has its own conditional distribution over decoder outcomes. We refer to $V_x$ as the \emph{response field} over these two coordinates. Formal definitions of the group action, the conditional decoder model, and the response field are collected in~\cref{app:preliminaries}.

\paragraph{Decoding before symmetrization.}
A \emph{decoding rule} $\mathcal D$ maps the conditional response distribution at one rewriting to an element of a real Hilbert space $\mathcal H$. We first describe this rule at the distribution level to isolate the cost of rewriting from the number of traces used to estimate it. The rule induces the \emph{decoded orbit field}
\begin{equation}
    F_x^{\mathcal D}(g)
    :=\mathcal D\!\left(p_{\theta,\tau}(\cdot\mid g\cdot x),
                         V_x(g,\cdot)\right)\in\mathcal H.
    \label{eq:decoded-orbit-field}
\end{equation}
For class prediction, one usually takes $\mathcal H=\mathbb R^C$. Three examples are
\begin{align}
    F_x^{\rm SC}(g)_c
    &:=\mathbb P_{\omega\mid g}\bigl(a_x(g,\omega)=c\bigr),
    \label{eq:decoder-sc}\\
    F_x^{\rm logit}(g)
    &:=\mathbb E_{\omega\mid g}\bigl[V_x(g,\omega)\bigr],
    \label{eq:decoder-logit}\\
    F_x^{\beta_D}(g)_c
    &:=\frac{\mathbb E_{\omega\mid g}
       \left[e^{\beta_D r_x(g,\omega)}
       \mathbf 1\{a_x(g,\omega)=c\}\right]}
       {\mathbb E_{\omega\mid g}
       \left[e^{\beta_D r_x(g,\omega)}\right]}.
    \label{eq:decoder-verifier}
\end{align}
Here $a_x(g,\omega)$ is the answer extracted from a trace and $r_x(g,\omega)$ is a verifier score. The first rule gives self consistency probabilities, the second averages logits, and the third is a verifier weighted vote. In~\cref{eq:decoder-verifier}, $\beta_D=0$ recovers self consistency and larger $\beta_D$ gives more weight to highly verified traces. This decoder inverse temperature is distinct from the generation temperature $\tau$ inside $p_{\theta,\tau}$. Our results hold simultaneously for all of these choices and for any other decoding rule whose output lies in a Hilbert space.

The \emph{fully symmetrized response} is
\begin{equation}
    q_x^{\mathcal D}:=\mathsf P_GF_x^{\mathcal D}
    =\frac{1}{|G|}\sum_{g\in G}F_x^{\mathcal D}(g).
    \label{eq:full-symmetrization}
\end{equation}
It is invariant to the starting rewriting, but it need not be constant across answer classes. For $\mathcal H=\mathbb R^C$, it remains a class score vector and can select a unique answer. Invariance is therefore imposed only in the rewriting coordinate, not in the answer coordinate. This is the core idea of model agnostic symmetrization: a frozen model is left unchanged, and task structure is imposed through external averaging.

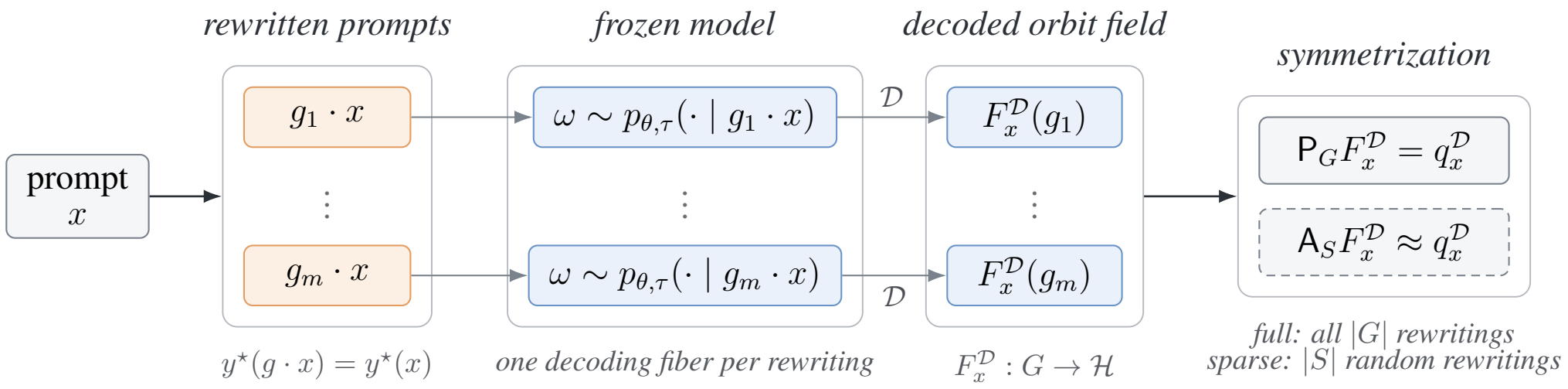
\begin{figure}[t]
    \centering
    \resizebox{\linewidth}{!}{%
        \input{figures/pipeline}
    }
    \caption{Decoding each rewriting produces the orbit field $F_x^{\mathcal D}$. Full group averaging gives the invariant target $q_x^{\mathcal D}$, while sparse averaging uses only the rewritings in $S$ to approximate it.}
    \label{fig:pipeline}
\end{figure}

Let $e\in G$ denote the identity rewriting, so that $e\cdot x=x$. Self consistency on the original prompt estimates $F_x^{\rm SC}(e)$, whereas full test time symmetrization returns $\mathsf P_GF_x^{\rm SC}$. Repeated decoding at $e\cdot x$ provides information only about the original representation. Querying $g\cdot x$ reveals the decoded response at another point on the orbit. Describing the decoding rule at the distribution level removes trace estimation error from the rewriting scaling law, but it does not erase this information difference.

\paragraph{A reusable symmetrization operator.}
A full average over the orbit requires $|G|$ forward passes. To reduce this cost, let
$S=\{\!\{s_1,\ldots,s_{|S|}\}\!\}$ be a multiset of group elements and define the sparse rewriting operator
\begin{equation}
    (\mathsf A_Sf)(h):=\frac{1}{|S|}\sum_{i=1}^{|S|} f(s_i h),
    \qquad h\in G.
    \label{eq:sparse-operator}
\end{equation}
The argument $h$ indexes the starting rewriting $h\cdot x$ from which the same inference procedure is invoked. The full group averaging operator is
\begin{equation}
    (\mathsf P_Gf)(h):=\frac{1}{|G|}\sum_{g\in G}f(gh),
    \label{eq:reynolds-projector}
\end{equation}
which is constant over $h$ and is the orthogonal projection onto the subspace of functions that are constant on $G$, as proved in~\cref{lem:group-projection}. It is important to distinguish the group coordinate from the answer coordinate. The invariant subspace consists of score fields that are constant as the starting representation $h$ changes. It does not require the scores of the $C$ answer classes to be equal. Indeed, if $f(h)=(f_1(h),\ldots,f_C(h))$, then
\begin{equation}
    (\mathsf P_Gf)_c(h)=\frac{1}{|G|}\sum_{g\in G}f_c(gh),
    \qquad c\in\mathcal C,
    \label{eq:coordinatewise-reynolds}
\end{equation}
is independent of $h$ but can still vary with $c$. Consequently, the invariant score vector $\mathsf P_Gf$ can select a unique answer through $\arg\max_c(\mathsf P_Gf)_c$. The projection therefore removes sensitivity to how the prompt is represented without removing the information needed to distinguish answer classes. If the original score field is already invariant, then $\mathsf P_Gf=f$.

Our goal is to derive a scaling law for the number $|S|$ of forward passes spent exploring $G$. The sampled multiset $S$ is chosen once and remains valid uniformly over frozen models, decoding rules, generation and verifier temperatures, decoder realizations, and output spaces, with error measured in mean square over starting representations. This separates the forward pass cost of rewriting from the statistical cost of estimating a chosen decoder and makes $S$ a reusable model agnostic symmetrization rule.

\subsection{Why decoder resampling alone is insufficient}
\label{sec:impossibility}

We show that drawing more reasoning traces from the original prompt cannot, in general, recover the decision obtained by averaging over rewritings. Two possible model behaviors can be identical on the original prompt but have different average behavior over its orbit. No decoding rule that observes only the original prompt can distinguish between them.

\begin{theorem}[Decoder resampling cannot replace rewriting]
\label{thm:two-worlds}
Let $G$ contain at least two elements and let $C\ge 2$. Consider any possibly randomized inference algorithm that is allowed to draw an arbitrary number of independent traces
\[
    \omega_t\sim p_{\theta,\tau}(\cdot\mid x)
\]
from the original prompt. The algorithm may use internal randomness together with all previous traces and scores to decide whether to draw another trace or to stop and output a class, but it never queries $g\cdot x$ for $g\ne e$. There exist two bounded response fields $V_x^{(0)}$ and $V_x^{(1)}$, sharing the same decoder distribution at $x$, such that
\[
    V_x^{(0)}(e,\omega)=V_x^{(1)}(e,\omega)
    \qquad\text{for every }\omega\in\Omega.
\]
Thus, the complete observation sequence of the algorithm has the same distribution in the two worlds, regardless of how many traces it draws. Nevertheless, the two worlds have different unique invariant decisions:
\[
    \arg\max_{c\in\mathcal C}(\mathsf P_GF_x^{(0)})_c
    \ne
    \arg\max_{c\in\mathcal C}(\mathsf P_GF_x^{(1)})_c,
\]
where $F_x^{(j)}(g):=\mathbb E_{\omega\mid g}[V_x^{(j)}(g,\omega)]$.
Consequently, if the two worlds are equally likely, every such algorithm has error at least $1/2$, regardless of its decoder budget.
\end{theorem}

The reason is simple. The algorithm receives exactly the same information in both worlds and must therefore behave in the same way, but the correct invariant decisions are different. It must fail in at least one world. Thus, additional decoding can reduce randomness at the original prompt, but it cannot reveal how the model behaves at unseen rewritings. The construction and proof are given in~\cref{app:proof-theorem-one}.

\section{Scaling Laws for Model-Agnostic Symmetrization}
\label{sec:results}

We now quantify how many rewritings are needed to approximate the full group average. The result is stated briefly here; its stronger operator and representation theoretic form appears in~\cref{thm:sparse-reynolds-full}.

\begin{theorem}[Sparse rewriting scaling law, informal]
\label{thm:sparse-reynolds}
Draw the rewritings in $S$ independently and uniformly from $G$. With probability at least $1-\delta$, the same set $S$ satisfies, for every real Hilbert space $\mathcal H$ and every $F\in L^2(G;\mathcal H)$,
\begin{equation}
    \|\mathsf A_SF-\mathsf P_GF\|_{L^2(G;\mathcal H)}
    \le \frac{8}{3}\sqrt{\frac{\log(|G|/\delta)}{|S|}}
    \|F-\mathsf P_GF\|_{L^2(G;\mathcal H)}.
    \label{eq:hilbert-spectral-bound}
\end{equation}
This guarantee is simultaneous over all such fields, not one field fixed in advance. Moreover, the $\log|G|$ dependence is optimal in general: for some finite groups, fewer than $\log_2|G|$ rewritings cannot achieve any uniform error below one.
\end{theorem}

Thus, $|S|=O(\varepsilon^{-2}(\log|G|+\log(1/\delta)))$ rewritings give relative $L^2$ error at most $\varepsilon$. The event depends only on $S$, which is why a single sampled set works for every model, decoder, trace realization, and answer space. In particular, there is no dependence on the number of answer classes. The complete theorem also shows that the same sample approximates averaging in every unitary representation. Its proof and the matching lower bound are given in~\cref{app:proof-sparse-averaging}. The argument uses random Cayley expansion and adapts the operator and averaging complexity ideas of \citet{alon1994random,tahmasebi2026augmentation,tahmasebi2026approximate}.

\subsection{A decoder agnostic scaling law}
\label{sec:main-scaling-law}

We now apply the same event after decoding. This order keeps the decoding rule arbitrary while leaving the rewriting rate unchanged.

\begin{corollary}[Decoder agnostic symmetrization]
\label{thm:decoder-agnostic}
On the event in~\cref{thm:sparse-reynolds}, simultaneously for every prompt, frozen model, generation temperature, decoding rule, and Hilbert valued output,
\begin{equation}
    \|\mathsf A_SF_x^{\mathcal D}-q_x^{\mathcal D}\|_{L^2(G;\mathcal H)}
    \le \frac{8}{3}\sqrt{\frac{\log(|G|/\delta)}{|S|}}
    \|F_x^{\mathcal D}-q_x^{\mathcal D}\|_{L^2(G;\mathcal H)}.
    \label{eq:decoder-scaling-law}
\end{equation}
The same statement holds for every realized randomized decoder.
\end{corollary}

There is no dependence on $C$, on the size of the trace space, or on a decoder covering number. A single draw of $S$ therefore works for self consistency, expected logits, every verifier temperature in~\cref{eq:decoder-verifier}, and any other decoding rule selected after $S$ is drawn. The guarantee is also pathwise: a finite randomized decoder produces a function on $G$, and the same event controls every seed and trace realization without a union bound over the trace space. It compares sparse and full group averaging of the same realized decoder field. Note that estimating a distribution level decoder from finitely many traces is a separate statistical question and does not alter the rewriting scaling law. The complete statement and proof are given in~\cref{app:proof-decoder-agnostic}; \cref{app:joint-sampling} explains why directly sampling joint pairs $(g,\omega)$ gives a different guarantee.

This universal control is what separates the two sources of diversity. The rewriting set is chosen without reference to the reasoning traces or to how they will be decoded. One may therefore use any trace sampling and decoding procedure at each rewriting, while the same event controls the resulting field. The rewriting budget $|S|$ and the trace budget can then be calibrated independently: our scaling law controls the former, while a decoder specific analysis controls the latter. No uniform estimation of every individual trace is required. An explicit decomposition of these two errors is given in~\cref{app:joint-sampling}.

This distinction also explains why we average over rewritings. If the goal were instead to find the single best rewriting, an isolated maximizer could require $\Theta(|G|)$ random queries to discover. The logarithmic law concerns approximate invariance under group averaging, not worst case search over the orbit.

\subsection{From symmetrization to prediction error}
\label{sec:decision-corollary}

The scaling law controls scores. We next translate it into prediction error. Let $X$ be drawn from the task distribution and take $\mathcal H=\mathbb R^C$. Fix any decoding rule and abbreviate $F_X=F_X^{\mathcal D}$ and $q_X=\mathsf P_GF_X$. Let
\[
    c_G(X):=\arg\max_{c\in\mathcal C}(q_X)_c,
    \qquad
    \widehat c_S(h\cdot X):=\arg\max_{c\in\mathcal C}(\mathsf A_SF_X)(h)_c,
\]
using a fixed tie breaking rule. Write $\Delta_X$ for the gap between the largest and second largest coordinates of $q_X$, with $\Delta_X=0$ at a tie, and define
\[
    R_G:=\mathbb P_X\bigl(c_G(X)\ne y^\star(X)\bigr),
    \qquad
    R_S:=\mathbb P_{X,h}\bigl(\widehat c_S(h\cdot X)\ne y^\star(X)\bigr),
\]
where $h\sim\Unif(G)$ independently of $X$.

\begin{corollary}[Prediction error decomposition]
\label{cor:margin}
On the event in~\cref{thm:sparse-reynolds}, for every $t>0$,
\begin{equation}
    R_S
    \le R_G
    +\mathbb P_X(\Delta_X\le t)
    +\frac{128}{9t^2}
      \frac{\log(|G|/\delta)}{|S|}
      \mathbb E_X\!\left[
      \|F_X-\mathsf P_GF_X\|_{L^2(G;\mathbb R^C)}^2
      \right].
    \label{eq:risk-decomposition}
\end{equation}
\end{corollary}

The three terms play different roles. The first, $R_G$, is the error of the fully symmetrized model, which uses all $|G|$ rewritings; it is a property of the model and the task, and no number of rewritings can reduce it. The second counts examples whose symmetrized scores have a small margin $\Delta_X$, where a small approximation error can change the decision. The third is the finite rewriting error, which decays as $\log|G|/|S|$ with no explicit dependence on the number of classes $C$; the threshold $t$ trades it against the second term. The bound thus separates three requirements. A large $|S|$ controls only the third term, and the decision can remain unstable when the margin is small. A large margin controls the second term, and even then a task can be hard for the model regardless of how it is written, in which case $R_G$ stays large. In particular, the result does not assume that full symmetrization is optimal: it separates this modeling error from the compute used to approximate it. The proof is given in~\cref{app:proof-margin}.

\input{sections/experiments}

\section{Conclusion}
\label{sec:conclusion}

We studied test-time symmetrization, which spends forward passes on rewritings of the input rather than on repeated decoding. Decoding alone cannot replace rewriting, whatever its budget, while one random set of $O(\log|G|)$ rewritings approximates the full group average simultaneously for every frozen model, decoding rule, realized decoder, and Hilbert valued output, with no dependence on the number of answer classes or the size of the reasoning space. An error decomposition separates the accuracy of the fully symmetrized model, its margin, and the finite rewriting cost. \rev{Experiments with four model families confirm the scaling law on next-token distributions and its consequences on reasoning traces. At the same number of forward passes, rewriting improves on resampling whenever the model's answer depends on how the input is written and the symmetrized decision is more accurate, across graph reasoning, logical deduction, and multiple choice with an equivariant answer, and it cannot help elsewhere.}

Our guarantees control invariance in mean square over starting representations rather than at every representation, and they concern the approximation of the fully symmetrized model rather than its accuracy, which depends on the model and the task. They also require the symmetry to be known and exact. \rev{When it is not, rewritings can be drawn from a kernel such as a paraphrasing model. The sparse average then estimates the kernel average at the same $|S|^{-1/2}$ rate, but the target is only meaningful up to the drift of the kernel, which adds to the error of \cref{cor:margin}, and nothing is uniform over an orbit. Our paraphrase experiment (\cref{app:exp-paraphrase}) shows that this substitute recovers part of the gain and that its quality is limited by the drift and the diversity of the paraphraser.} Natural next steps are \rev{a theory for such approximate symmetries,} symmetrization inside reasoning traces rather than only at the answer, and further exact symmetries such as identifier renaming.

\subsection*{AI use statement}

Large language model tools (ChatGPT 5.6 Sol and Claude Fable 5.1) were used primarily for copyediting, including improvements to grammar, wording, clarity, and \LaTeX{} presentation. The
research direction, core ideas, theoretical results, and experimental design
are original contributions of the authors. The authors carefully reviewed and
verified all technical content and take full responsibility for the final
manuscript.

\section*{Acknowledgments}
BT and MW were supported by NSF award DMS-2406905, Coefficient Giving, and the Harvard Dean's Fund for Promising Scholarship. MW acknowledges partial support from an Alfred P. Sloan Research Fellowship in Mathematics and the AI2050 program at Schmidt Sciences (Grant G-25-69786).

\bibliography{ref}
\bibliographystyle{iclr2027_conference}

\clearpage
\appendix
\addtocontents{toc}{\protect\setcounter{tocdepth}{2}}
\begingroup
\renewcommand{\contentsname}{Appendix Contents}
\tableofcontents
\endgroup

\section{Additional Related Work}
\label{app:related}

\paragraph{Test-time augmentation.}
Averaging predictions over transformed inputs at inference is an old idea in vision, going back to multi-crop evaluation \citep{krizhevsky2012imagenet,simonyan2015vgg}. Later work studies how to aggregate the augmented predictions \citep{shanmugam2021aggregation}, when averaging helps \citep{kimura2021understanding}, and how it compares with ensembles \citep{ashukha2020pitfalls,breiman1996bagging,lakshminarayanan2017ensembles}. Our setting differs in that the transformations form a group, so the cost can be measured against $|G|$, and the predictor is itself random.

\paragraph{Model-agnostic symmetrization.}
Symmetry is usually built into the architecture \citep{cohen2016group,bronstein2021geometric,zaheer2017deepsets,maron2019invariant}. When this is not possible, it can be imposed from outside by averaging over the group: Janossy pooling \citep{murphy2019janossy}, frame averaging \citep{puny2022frame}, canonicalization \citep{kaba2023equivariance,mondal2023equivariant}, probabilistic symmetrization \citep{kim2023probabilistic}, and equi-tuning of pretrained models \citep{basu2023equituning} are examples\rev{, and \citet{tahmasebi2025canonicalization} compare the generalization of canonicalization with that of group averaging}. Related work measures how equivariant a trained model is \citep{gruver2023lie}, learns invariances from data \citep{benton2020learning}, or studies approximate symmetries \citep{petrache2023approximation}. These methods generally alter model construction, training, or internal representations. Our model is frozen and accessed only through forward passes, so the cost of symmetrization is counted in queries.

\paragraph{Theory of invariance.}
Invariance is known to help at training time: data augmentation reduces variance \citep{chen2020grouptheoretic,lyle2020benefits}, equivariant models can generalize strictly better \citep{elesedy2021provably}, and the gain has been quantified for kernel models \citep{bietti2021sample,mei2021learning,tahmasebi2023exact}\rev{, for divergence estimation \citep{tahmasebi2024divergences},} and measured experimentally \citep{adly2026symmetry}; see also \citet{tahmasebi2026augmentation} for a Fourier view of augmentation. Most relevant to us, \citet{tahmasebi2026approximate} show in their averaging-complexity framework that exact invariance can require all $|G|$ group elements while approximate invariance needs only $\Theta(\log|G|)$ random ones. Our \cref{thm:sparse-reynolds} is of the same kind and relies on random Cayley graphs being expanders \citep{alon1994random,landau2004random} and on matrix concentration \citep{tropp2012user,tropp2015introduction}. We extend this from deterministic functions to stochastic frozen models, make the guarantee uniform, combine it with decoding, and connect it to decisions through the margin. \rev{More precisely, \citet{tahmasebi2026approximate} bound the averaging complexity of a fixed, deterministic function class. A single sparse set of group elements drawn once has also been used at training time, to approximate the group-averaged gradient uniformly over all parameters \citep{tahmasebi2026sparse}. What is new here is the separation of the two sources of randomness (decoding fibers and rewriting orbits) and the impossibility result that follows from it (\cref{thm:two-worlds}), the uniformity of a single draw over every decoding rule and every realized trace without a covering-number penalty (\cref{thm:decoder-agnostic}, through \cref{lem:hilbert-tensorization,lem:regular-universal}), the margin decomposition that turns the approximation into a prediction error (\cref{cor:margin}), and the question of allocating a test-time budget between the two sources. Synthetic graphs with random node identifiers, as in our experiments, are the standard setting for isolating what language models learn about structure \citep{bachmann2024pitfalls,saparov2025transformers,frydenlund2025graph,nagarajan2025roll,noroozizadeh2026geometric,wang2026minimal}.}

\section{Mathematical Preliminaries}
\label{app:preliminaries}

This appendix collects the definitions, notation, and elementary facts used throughout the paper. All expectations of vector valued functions are understood coordinatewise.

\subsection{Finite groups, actions, and orbits}

\begin{definition}[Finite group]
A \emph{finite group} $G$ is a finite set equipped with an associative product $(g,h)\mapsto gh$, an identity element $e\in G$, and an inverse $g^{-1}\in G$ for every $g\in G$. Thus, $eg=ge=g$ and $g^{-1}g=gg^{-1}=e$.
\end{definition}

\begin{definition}[Group action and orbit]
A left \emph{action} of $G$ on a set $\mathcal X$ is a map $(g,x)\mapsto g\cdot x$ satisfying
\[
    e\cdot x=x,
    \qquad
    (gh)\cdot x=g\cdot(h\cdot x)
\]
for all $g,h\in G$ and $x\in\mathcal X$. The \emph{orbit} of $x$ is
\[
    G\cdot x=\{g\cdot x:g\in G\}.
\]
Elements of the same orbit are the rewritings that the task regards as semantically equivalent.
\end{definition}

\begin{definition}[Invariant and equivariant targets]
Let $y^\star:\mathcal X\to\mathcal C$ be the target function. It is \emph{invariant} if
\[
    y^\star(g\cdot x)=y^\star(x)
\]
for every $g\in G$ and $x\in\mathcal X$. More generally, suppose that each $g\in G$ induces a permutation $\rho(g):\mathcal C\to\mathcal C$ of the answer classes, with $\rho(gh)=\rho(g)\rho(h)$. The target is \emph{equivariant} if
\[
    y^\star(g\cdot x)=\rho(g)y^\star(x).
\]
In the equivariant case, applying $\rho(g)^{-1}$ to the output scores aligns them with the original answer coordinates and reduces the problem to the invariant setting.
\end{definition}

\subsection{Conditional decoding and response fields}

Equip $\mathcal X$ with a sigma algebra $\mathcal B_\mathcal X$, and let $(\Omega,\mathcal B_\Omega)$ be the measurable space of reasoning traces. A frozen model specifies, for every prompt $x\in\mathcal X$, a probability measure
\[
    p_{\theta,\tau}(\cdot\mid x)
\]
on $\Omega$. Equivalently, $p_{\theta,\tau}$ is a conditional probability kernel from prompts to reasoning traces, where $\tau$ may specify a generation temperature. Querying the model on $g\cdot x$ means drawing
\[
    \omega\sim p_{\theta,\tau}(\cdot\mid g\cdot x).
\]
We assume throughout that the maps below are measurable and square integrable whenever an $L^2$ norm is used.

\begin{definition}[Response field and decoding fiber]
For a fixed semantic instance $x$, a \emph{response field} is a measurable function
\[
    V_x:G\times\Omega\to\mathcal Z.
\]
The value $V_x(g,\omega)$ is the response representation obtained from trace $\omega$ at rewriting $g\cdot x$. It may contain an answer, logits, a verifier score, or the trace itself. The pair
\[
    \bigl(\Omega,p_{\theta,\tau}(\cdot\mid g\cdot x)\bigr)
\]
is the \emph{decoding fiber} at $g\cdot x$.
\end{definition}

The joint rewriting and decoding experiment is
\[
    g\sim\Unif(G),
    \qquad
    \omega\mid g\sim p_{\theta,\tau}(\cdot\mid g\cdot x).
\]
When $V_x$ is vector valued and integrable, its joint expectation is
\[
    \mathbb E_{g,\omega\mid g}[V_x(g,\omega)]
    =\frac{1}{|G|}\sum_{g\in G}
      \int_\Omega V_x(g,\omega)
      p_{\theta,\tau}(d\omega\mid g\cdot x).
\]

\begin{definition}[Distribution level decoding rule]
Let $\mathcal H$ be a real Hilbert space. A distribution level decoding rule is any measurable map $\mathcal D$ from the conditional response distribution at a rewriting to an element of $\mathcal H$. Its decoded orbit field is
\[
    F_x^{\mathcal D}(g)
    =\mathcal D\!\left(p_{\theta,\tau}(\cdot\mid g\cdot x),
                        V_x(g,\cdot)\right).
\]
No linearity or finite dimensionality is assumed. A randomized finite decoder with seed $\xi$ similarly induces a realized field $\widehat F_{x,\xi}:G\to\mathcal H$.
\end{definition}

\subsection{Functions on the group and invariant projection}

For a real Hilbert space $\mathcal H$, equip $L^2(G;\mathcal H)$ with the uniform inner product
\[
    \langle f,q\rangle_{L^2(G)}
    :=\frac{1}{|G|}\sum_{h\in G}
      \langle f(h),q(h)\rangle_{\mathcal H}.
\]

\begin{lemma}[Full group averaging]
\label[lemma]{lem:group-projection}
For $f\in L^2(G;\mathcal H)$, define
\[
    (\mathsf P_Gf)(h)
    :=\frac{1}{|G|}\sum_{g\in G}f(gh).
\]
Then $\mathsf P_Gf$ is constant in $h$, and $\mathsf P_G$ is the orthogonal projection onto the subspace of functions that are constant on $G$.
\end{lemma}

\begin{proof}
For fixed $h\in G$, the map $g\mapsto gh$ is a bijection of $G$. Hence
\[
    (\mathsf P_Gf)(h)
    =\frac{1}{|G|}\sum_{u\in G}f(u),
\]
which does not depend on $h$. Thus, the image of $\mathsf P_G$ consists of constant functions, and $\mathsf P_Gq=q$ for every constant function $q$. Moreover, $f-\mathsf P_Gf$ has mean zero and is therefore orthogonal to every constant function. This proves the claim.
\end{proof}

\subsection{Translation operators and matrix concentration}

For a scalar function $f\in L^2(G)$ and $s\in G$, define the translation operator
\begin{equation}
    (\mathsf T_s f)(h):=f(sh),
    \qquad h\in G.
    \label{eq:translation-operator}
\end{equation}
Because $h\mapsto sh$ is a permutation of $G$, $\mathsf T_s$ is unitary under the uniform measure. In this notation,
\begin{equation}
    \mathsf A_S=\frac{1}{|S|}\sum_{i=1}^{|S|}\mathsf T_{s_i},
    \qquad
    \mathsf P_G=\mathbb E_{s\sim\Unif(G)}[\mathsf T_s].
    \label{eq:operators-as-translations}
\end{equation}
The operator norm used in~\cref{thm:sparse-reynolds-full} is
\begin{equation}
    \|\mathsf T\|_{L^2(G)\to L^2(G)}
    :=\sup_{f\ne0}\frac{\|\mathsf T f\|_{L^2(G)}}{\|f\|_{L^2(G)}}.
    \label{eq:l2-operator-norm}
\end{equation}
Thus, an operator norm bound is one event that controls the approximation error for every function $f\in L^2(G)$ simultaneously.

We use the following standard matrix Bernstein inequality~\citep{tropp2012user}. It is stated for operators so that it applies directly to translations on $L^2(G)$.

\begin{lemma}[Matrix Bernstein inequality]
\label[lemma]{lem:matrix-bernstein}
Let $X_1,\ldots,X_m$ be independent, mean zero linear operators on a $d$ dimensional Hilbert space. Suppose that $\|X_i\|_{\mathrm{op}}\le R$ almost surely, and define
\[
    \sigma^2:=\max\left\{
      \left\|\sum_{i=1}^m\mathbb E[X_iX_i^*]\right\|_{\mathrm{op}},
      \left\|\sum_{i=1}^m\mathbb E[X_i^*X_i]\right\|_{\mathrm{op}}
    \right\}.
\]
Then, for every $t\ge0$,
\[
    \mathbb P\left(\left\|\sum_{i=1}^mX_i\right\|_{\mathrm{op}}\ge t\right)
    \le 2d\exp\left(
      -\frac{t^2}{2\sigma^2+2Rt/3}
    \right).
\]
\end{lemma}

The nonsymmetric form above follows from the self-adjoint matrix Bernstein inequality by applying it to the self-adjoint dilation of each $X_i$.

\subsection{Hilbert valued averaging and unitary representations}

We collect the representation theoretic facts behind the universality in~\cref{thm:sparse-reynolds-full}. We state them over complex Hilbert spaces, where the standard Fourier decomposition is most transparent. A real representation is covered by passing to its complexification, which preserves operator norms.

\begin{definition}[Unitary representation and invariant subspace]
A \emph{unitary representation} of $G$ on a Hilbert space $\mathcal H$ is a map $\pi:G\to\mathcal U(\mathcal H)$ satisfying
\[
    \pi(gh)=\pi(g)\pi(h),
    \qquad \pi(e)=I_{\mathcal H}.
\]
Its invariant subspace is
\[
    \mathcal H^G
    :=\{v\in\mathcal H:\pi(g)v=v\text{ for every }g\in G\}.
\]
A nonzero representation is \emph{irreducible} if its only closed invariant subspaces are $\{0\}$ and the whole space.
\end{definition}

The \emph{left regular representation} $\lambda$ acts on $L^2(G;\mathbb C)$ by
\[
    (\lambda(g)f)(h):=f(g^{-1}h).
\]
Our translation $\mathsf T_gf(h)=f(gh)$ equals $\lambda(g^{-1})$. This inverse changes neither the relevant operator norm nor the distribution of a uniform group sample.

Let $\widehat G$ denote a complete set of inequivalent irreducible unitary representations, and let $d_\rho$ be the dimension of $\rho\in\widehat G$. The finite group Fourier decomposition gives
\begin{equation}
    L^2(G;\mathbb C)
    \cong
    \bigoplus_{\rho\in\widehat G}
       \mathbb C^{d_\rho}\otimes\mathbb C^{d_\rho},
    \qquad
    \lambda(g)
    \cong
    \bigoplus_{\rho\in\widehat G}
       \rho(g)\otimes I_{d_\rho}.
    \label{eq:regular-fourier-decomposition}
\end{equation}
Thus, every irreducible representation occurs inside the regular representation, with multiplicity equal to its dimension. More generally, every unitary representation of a finite group is an orthogonal direct sum of irreducible representations. These are the finite group forms of Maschke's theorem and the Peter Weyl decomposition.

The Hilbert valued space $L^2(G;\mathcal H)$ is naturally the Hilbert tensor product $L^2(G)\otimes\mathcal H$. A scalar operator $T$ on $L^2(G)$ acts on it as $T\otimes I_{\mathcal H}$.

\begin{lemma}[Hilbert valued tensorization]
\label[lemma]{lem:hilbert-tensorization}
For every bounded operator $T:L^2(G)\to L^2(G)$ and every real Hilbert space $\mathcal H$,
\[
    \|T\otimes I_{\mathcal H}\|_{L^2(G;\mathcal H)\to L^2(G;\mathcal H)}
    =\|T\|_{L^2(G)\to L^2(G)}.
\]
Consequently, if $T\mathsf P_G=\mathsf P_GT=0$, then
\[
    \|TF\|_{L^2(G;\mathcal H)}
    \le \|T\|_{\mathrm{op}}
       \|F-\mathsf P_GF\|_{L^2(G;\mathcal H)}.
\]
\end{lemma}

\begin{proof}
Choose an orthonormal basis $(u_j)_j$ for the closed span of the finite set $\{F(g):g\in G\}$. Writing $F=\sum_j f_j u_j$ and using Parseval's identity gives
\[
    \|(T\otimes I_{\mathcal H})F\|^2
    =\sum_j\|Tf_j\|_{L^2(G)}^2
    \le\|T\|_{\mathrm{op}}^2\sum_j\|f_j\|_{L^2(G)}^2.
\]
This proves the upper bound, and testing on $F=f u$ for a unit vector $u\in\mathcal H$ proves equality. The final claim follows from $TF=T(F-\mathsf P_GF)$.
\end{proof}

\begin{lemma}[Group averaging is an invariant projection]
\label[lemma]{lem:representation-projection}
For a unitary representation $\pi:G\to\mathcal U(\mathcal H)$, the operator
\[
    \mathsf P_\pi:=\frac{1}{|G|}\sum_{g\in G}\pi(g)
\]
is the orthogonal projection onto $\mathcal H^G$.
\end{lemma}

\begin{proof}
For any $a\in G$, a change of variables gives
\[
    \pi(a)\mathsf P_\pi
    =\frac{1}{|G|}\sum_{g\in G}\pi(ag)
    =\mathsf P_\pi.
\]
Hence the image of $\mathsf P_\pi$ lies in $\mathcal H^G$. Conversely, $\mathsf P_\pi v=v$ for every $v\in\mathcal H^G$. Moreover,
\[
    \mathsf P_\pi^*
    =\frac{1}{|G|}\sum_{g\in G}\pi(g^{-1})
    =\mathsf P_\pi,
\]
and the same change of variables gives $\mathsf P_\pi^2=\mathsf P_\pi$. It is therefore the orthogonal projection onto $\mathcal H^G$.
\end{proof}

\begin{lemma}[The regular representation controls every representation]
\label[lemma]{lem:regular-universal}
Let $\pi:G\to\mathcal U(\mathcal H)$ be a unitary representation of a finite group, and let
\[
    \mathsf P_\pi:=\frac{1}{|G|}\sum_{g\in G}\pi(g).
\]
For every multiset $S\subseteq G$,
\[
    \left\|\frac{1}{|S|}\sum_{s\in S}\pi(s)-\mathsf P_\pi\right\|_{\mathrm{op}}
    \le \|\mathsf A_S-\mathsf P_G\|_{L^2(G)\to L^2(G)}.
\]
Moreover, $\mathsf P_\pi$ is the orthogonal projection onto the vectors fixed by every $\pi(g)$.
\end{lemma}

\begin{proof}
Write $m=|S|$ and define
\[
    B_\pi:=\frac{1}{m}\sum_{i=1}^m\pi(s_i)-\mathsf P_\pi.
\]
Decompose $\pi$ into irreducibles. On an isotypic component associated with $\rho\in\widehat G$, the operator $B_\pi$ acts as
\[
    \left(\frac{1}{m}\sum_{i=1}^m\rho(s_i)-\mathsf P_\rho\right)
    \otimes I,
\]
so $\|B_\pi\|_{\mathrm{op}}$ is the supremum of the norms of these irreducible blocks. By~\cref{eq:regular-fourier-decomposition}, every such block occurs in the regular representation.

Our convention uses $\mathsf T_s=\lambda(s^{-1})$. The corresponding regular block contains
\[
    \frac{1}{m}\sum_{i=1}^m\rho(s_i^{-1})-\mathsf P_\rho
    =\left(\frac{1}{m}\sum_{i=1}^m\rho(s_i)-\mathsf P_\rho\right)^*.
\]
Taking an adjoint does not change the operator norm. Therefore, each irreducible block of $B_\pi$ has norm at most
$\|\mathsf A_S-\mathsf P_G\|_{L^2(G)\to L^2(G)}$, and taking the supremum over the blocks proves the inequality. The final statement follows from~\cref{lem:representation-projection}.
\end{proof}

\subsection{Margins and decision disagreement}
\label{app:decision-preliminaries}

For a score vector $q\in\mathbb R^C$ with a unique maximizer $c^*$, define its margin
\[
    \Delta(q):=q_{c^*}-\max_{c\ne c^*}q_c.
\]
We use any fixed rule to break ties in approximate score vectors.

\begin{lemma}[A changed decision requires score error]
\label[lemma]{lem:margin-error}
Let $q,v\in\mathbb R^C$, suppose that $q$ has unique maximizer $c^*$, and let $\Delta(q)>0$ be its margin. If $\arg\max_c v_c\ne c^*$, then
\[
    \|v-q\|_2^2\ge\frac{\Delta(q)^2}{2}.
\]
\end{lemma}

\begin{proof}
Let $c\ne c^*$ be the class selected by $v$. Since $v_c\ge v_{c^*}$,
\[
    \Delta(q)
    \le q_{c^*}-q_c
    \le |v_{c^*}-q_{c^*}|+|v_c-q_c|
    \le\sqrt2\,\|v-q\|_2.
\]
Squaring both sides proves the claim.
\end{proof}

\section{On Joint Rewriting and Decoding Sampling}
\label{app:joint-sampling}

A natural alternative is to sample pairs
\[
    g_i\sim\Unif(G),
    \qquad
    \omega_i\sim p_{\theta,\tau}(\cdot\mid g_i\cdot x),
\]
and aggregate the resulting responses directly. This is a valid inference procedure, but it answers a different statistical question from the operator approximation studied in this paper. We explain the distinction here.

\paragraph{Joint sampling is exact for linear decoding rules.}
Suppose that a distribution level decoding rule has the linear form
\[
    F_x^{\mathcal D}(g)
    =\mathbb E_{\omega\mid g}\bigl[\psi_x(g,\omega)\bigr]
\]
for some Hilbert valued response $\psi_x$. Then the fully symmetrized response satisfies
\begin{equation}
    q_x^{\mathcal D}
    =\mathbb E_{g\sim\Unif(G)}
       \mathbb E_{\omega\mid g}[\psi_x(g,\omega)]
    =\mathbb E_{g,\omega\mid g}[\psi_x(g,\omega)].
    \label{eq:joint-linear-identity}
\end{equation}
Consequently, independent joint samples give an unbiased Monte Carlo estimator of $q_x^{\mathcal D}$. Self consistency with one hot vote vectors and expected logits are examples. For one fixed instance, one fixed decoder, and one fixed starting representation, standard concentration may therefore be the simplest analysis.

\paragraph{Nonlinear decoding changes the target quantity.}
The equality in~\cref{eq:joint-linear-identity} does not hold for a general decoder. The quantity studied in this paper is
\begin{equation}
    q_x^{\mathcal D}
    =\mathbb E_g\bigl[\mathcal D(P_{x,g})\bigr],
    \qquad
    P_{x,g}:=\operatorname{Law}\bigl(V_x(g,\omega)\mid g\bigr),
    \label{eq:decode-then-symmetrize}
\end{equation}
which first decodes each conditional trace distribution and then removes rewriting sensitivity. Pooling joint samples generally corresponds instead to applying a decoder after mixing the conditional distributions. These operations need not commute:
\[
    \mathbb E_g\bigl[\mathcal D(P_{x,g})\bigr]
    \ne
    \mathcal D\!\left(\mathbb E_g P_{x,g}\right).
\]
For example, the verifier weighted decoder in~\cref{eq:decoder-verifier} averages the per rewriting ratios
\[
    \mathbb E_g\left[
    \frac{\mathbb E_{\omega\mid g}
      [e^{\beta_Dr_x(g,\omega)}\mathbf 1\{a_x(g,\omega)=c\}]}
    {\mathbb E_{\omega\mid g}[e^{\beta_Dr_x(g,\omega)}]}
    \right].
\]
Pooling all pairs before normalization instead produces a ratio of joint expectations. The two weight rewritings differently and can select different answers.

\paragraph{Pointwise estimation is not universal operator approximation.}
Even in the linear case, joint Monte Carlo estimates one chosen quantity $q_x^{\mathcal D}$ for one response field. Our goal is stronger: draw $S$ once so that the induced group operator approximates $\mathsf P_G$ for every response field simultaneously. This operator event remains valid after the model, decoder, generation temperature, verifier temperature, output dimension, and decoder realization are chosen. It also controls mean square sensitivity to the starting rewriting. A concentration bound for one joint expectation does not provide this reusable guarantee. Indeed, the lower bound in~\cref{thm:sparse-reynolds-full} shows that uniform approximation over all orbit functions requires the $\log|G|$ dependence that can be absent from a pointwise Monte Carlo bound.

\paragraph{What happens with finite trace sampling.}
The distribution level formulation isolates the cost of exploring the orbit. In practice, a finite decoder is run at each queried rewriting, producing a realized field $\widehat F_{x,\xi}$. The bound in~\cref{eq:pathwise-decoder-law} then compares sparse and full group averaging of that same realized field, simultaneously for every seed and trace configuration. Any difference between $\widehat F_{x,\xi}$ and the desired field $F_x^{\mathcal D}$ is a decoder estimation problem and depends on the chosen decoding algorithm. Keeping this term separate makes clear which error is reduced by more rewritings and which is reduced by more traces per rewriting.

The separation has a direct error decomposition. Since $\mathsf A_S$ is an average of unitary translation operators, $\|\mathsf A_S\|_{\mathrm{op}}\le 1$. On the event in~\cref{thm:sparse-reynolds-full},
\begin{align}
    \|\mathsf A_S\widehat F_{x,\xi}-q_x^{\mathcal D}\|_{L^2(G;\mathcal H)}
    &\le
    \|\widehat F_{x,\xi}-F_x^{\mathcal D}\|_{L^2(G;\mathcal H)}
    \notag\\
    &\quad+
    \frac{8}{3}\sqrt{\frac{\log(|G|/\delta)}{|S|}}
    \|F_x^{\mathcal D}-\mathsf P_GF_x^{\mathcal D}\|_{L^2(G;\mathcal H)}.
    \label{eq:finite-decoder-decomposition}
\end{align}
The first term is the finite decoding error and the second is the rewriting approximation error. They can be controlled independently and then combined through~\cref{eq:finite-decoder-decomposition}. In particular, the first term requires control only of the aggregate decoded field in $L^2(G;\mathcal H)$, not uniform approximation over the trace space $\Omega$.

For example, consider self consistency with $n_D$ independent traces at each rewriting. If $F_x^{\rm SC}(g)$ is the class probability vector and $\widehat F_x^{\rm SC}(g)$ is its empirical frequency vector, then
\begin{equation}
    \mathbb E\!\left[
      \|\widehat F_x^{\rm SC}-F_x^{\rm SC}\|_{L^2(G;\mathbb R^C)}^2
    \right]
    =\frac{1}{n_D}\,
      \mathbb E_{g\sim\Unif(G)}
      \left[1-\|F_x^{\rm SC}(g)\|_2^2\right]
    \le \frac{1}{n_D}.
    \label{eq:self-consistency-decoder-error}
\end{equation}
Thus, even this decoder term need not carry an explicit dependence on the number of classes. Averaged logits admit the usual Hilbert valued concentration bounds under a norm or tail assumption. Verifier weighted decoding additionally requires control of the weights and of its normalization denominator. These assumptions and rates belong to the selected decoder, whereas the rewriting rate is unchanged.

Thus, joint sampling is not ruled out, and it is fully appropriate for estimating a fixed linear joint expectation. It is not the object of our main scaling law because it can change the target for nonlinear decoders and, even when it does not, it does not yield the universal model agnostic symmetrization operator that is the focus of this paper.

\section{Proof of \texorpdfstring{Theorem~\ref{thm:two-worlds}}{Theorem 1}}
\label{app:proof-theorem-one}

\begin{proof}[Proof of Theorem~\ref{thm:two-worlds}]
Choose two distinct classes $c_0,c_1\in\mathcal C$, and let $u_{c_0},u_{c_1}\in\mathbb R^C$ be their standard basis vectors. Use the same decoder distribution in both worlds and define response fields that do not depend on $\omega$:
\[
    V_x^{(j)}(g,\omega)
    :=
    \begin{cases}
        0, & g=e,\\
        u_{c_j}, & g\ne e,
    \end{cases}
    \qquad j\in\{0,1\}.
\]
Both response fields take values in $[0,1]^C$. At the original prompt,
\[
    V_x^{(0)}(e,\omega)
    =V_x^{(1)}(e,\omega)=0
\]
for every $\omega$. Since the decoder distribution at $x$ is also the same, every trace and score observed by an algorithm that queries only $x$ has the same distribution in both worlds. This remains true under adaptive sampling because, conditional on any previous transcript, the next observation has the same conditional distribution in the two worlds. For a randomized algorithm, couple its internal random seed to be identical in both worlds. The complete transcript, including the internal randomness, then has the same distribution in both worlds.

The response fields are independent of $\omega$, so decoder averaging leaves them unchanged. Therefore,
\[
    \mathsf P_GF_x^{(j)}
    =\frac{|G|-1}{|G|}u_{c_j}.
\]
The unique invariant decision is $c_j$ in world $j$, and these decisions are different.

Let $Q$ be the common output distribution of the algorithm in the two worlds. Under a uniform prior on the worlds, its success probability is
\[
    \frac{1}{2}Q(c_0)+\frac{1}{2}Q(c_1)
    \le\frac{1}{2},
\]
because $c_0\ne c_1$. Hence its error probability is at least $1/2$, regardless of the number of decoder samples.
\end{proof}

\section{Proof of \texorpdfstring{Theorem~\ref{thm:sparse-reynolds}}{Theorem 2}}
\label{app:proof-sparse-averaging}

The proof adapts the operator concentration viewpoint used for partial group averaging in~\citet{tahmasebi2026augmentation} and the averaging complexity ideas of~\citet{tahmasebi2026approximate}. Here these ideas are applied to test time rewriting operators acting on frozen model response fields.

We first state the complete version of~\cref{thm:sparse-reynolds}. In addition to the Hilbert valued bound used in the main text, it records the underlying operator estimate and its consequence for every unitary representation of $G$.

\begin{theorem}[Universal sparse group averaging]
\label{thm:sparse-reynolds-full}
Let $S$ contain $|S|$ independent samples from $\Unif(G)$ and let $\delta\in(0,1)$. With probability at least $1-\delta$,
\begin{equation}
    \|\mathsf A_S-\mathsf P_G\|_{L^2(G)\to L^2(G)}
    \le
    \frac{8}{3}\sqrt{\frac{\log(|G|/\delta)}{|S|}}.
    \label{eq:full-spectral-bound}
\end{equation}
On the same event, for every real Hilbert space $\mathcal H$ and every $F\in L^2(G;\mathcal H)$,
\begin{equation}
    \|\mathsf A_SF-\mathsf P_GF\|_{L^2(G;\mathcal H)}
    \le
    \frac{8}{3}\sqrt{\frac{\log(|G|/\delta)}{|S|}}
    \|F-\mathsf P_GF\|_{L^2(G;\mathcal H)}.
    \label{eq:hilbert-spectral-bound-full}
\end{equation}
Moreover, for every unitary representation $\pi:G\to\mathcal U(\mathcal H)$, if $\mathsf P_\pi$ denotes the orthogonal projection onto its invariant vectors, then
\begin{equation}
    \left\|\frac{1}{|S|}\sum_{i=1}^{|S|}\pi(s_i)-\mathsf P_\pi
    \right\|_{\mathrm{op}}
    \le
    \frac{8}{3}\sqrt{\frac{\log(|G|/\delta)}{|S|}}.
    \label{eq:all-representations}
\end{equation}
Finally, the logarithmic dependence is necessary in general: for $G=(\mathbb Z/2\mathbb Z)^n$, every multiset with fewer than $n=\log_2|G|$ elements has operator error equal to one.
\end{theorem}

\begin{proof}[Proof of \cref{thm:sparse-reynolds-full}]
Write $m=|S|$ and $N=|G|$. The case $N=1$ is immediate, so assume $N\ge2$. For each sampled element $s_i$, let
\[
    X_i:=\mathsf T_{s_i}-\mathsf P_G.
\]
By~\cref{eq:operators-as-translations}, the operators $X_i$ are independent and have mean zero. Translation preserves constant functions and preserves the uniform mean, so
\[
    \mathsf T_s\mathsf P_G=\mathsf P_G\mathsf T_s=\mathsf P_G
    \qquad\text{for every }s\in G.
\]
Since $\mathsf T_s$ is unitary and $\mathsf P_G$ is an orthogonal projection, it follows that
\[
    X_iX_i^*=X_i^*X_i=I-\mathsf P_G,
    \qquad
    \|X_i\|_{\mathrm{op}}=1.
\]
Consequently, the variance parameter in~\cref{lem:matrix-bernstein} satisfies
\[
    \sigma^2
    =m\|I-\mathsf P_G\|_{\mathrm{op}}
    =m.
\]
Applying~\cref{lem:matrix-bernstein} on the $N$ dimensional space $L^2(G)$ gives, for every $u\ge0$,
\[
    \mathbb P\left(
      \left\|\sum_{i=1}^mX_i\right\|_{\mathrm{op}}\ge u
    \right)
    \le
    2N\exp\left(-\frac{u^2}{2m+2u/3}\right).
\]
Equivalently, for every $t\ge0$,
\begin{equation}
    \mathbb P\left(
      \|\mathsf A_S-\mathsf P_G\|_{L^2(G)\to L^2(G)}\ge t
    \right)
    \le
    2N\exp\left(-\frac{mt^2}{2+2t/3}\right).
    \label{eq:sparse-averaging-tail}
\end{equation}
Moreover, $\|\mathsf A_S-\mathsf P_G\|_{\mathrm{op}}\le1$, since on constant functions the difference vanishes and on their orthogonal complement it is an average of unitary operators. If
\[
    \frac{8}{3}\sqrt{\frac{\log(N/\delta)}{m}}\ge1,
\]
the claimed bound is therefore deterministic. Otherwise, set
\[
    t=\frac{8}{3}\sqrt{\frac{\log(N/\delta)}{m}}<1.
\]
Since $2+2t/3\le8/3$,~\cref{eq:sparse-averaging-tail} gives
\[
    \mathbb P\left(
      \|\mathsf A_S-\mathsf P_G\|_{\mathrm{op}}\ge t
    \right)
    \le2N\left(\frac{N}{\delta}\right)^{-8/3}
    \le\delta,
\]
where the last inequality uses $N/\delta\ge2$. This proves~\cref{eq:full-spectral-bound} with the explicit constant $8/3$.

Set $T=\mathsf A_S-\mathsf P_G$. Since $T\mathsf P_G=\mathsf P_GT=0$, \cref{lem:hilbert-tensorization} gives~\cref{eq:hilbert-spectral-bound-full} for every Hilbert space and every Hilbert valued field on the same event. Applying \cref{lem:regular-universal} gives~\cref{eq:all-representations} for every unitary representation, again on the same event. No further probability bound or union bound is required.

It remains to prove the worst case necessity of $\log|G|$. Let
$G=(\mathbb Z/2\mathbb Z)^n$, so that $|G|=2^n$, and suppose that $m<n$. The sampled elements $s_1,\ldots,s_m$ span a proper subspace of $G$ over $\mathbb Z/2\mathbb Z$. Hence there is a nonzero $a\in G$ such that
\[
    \langle a,s_i\rangle=0\pmod 2
    \qquad\text{for every }i.
\]
Define the nonconstant character $\chi_a(h)=(-1)^{\langle a,h\rangle}$. It has uniform mean zero, so $\mathsf P_G\chi_a=0$. On the other hand,
\[
    (\mathsf T_{s_i}\chi_a)(h)
    =\chi_a(s_i+h)
    =\chi_a(s_i)\chi_a(h)
    =\chi_a(h),
\]
and therefore $\mathsf A_S\chi_a=\chi_a$. Thus,
\[
    \|\mathsf A_S-\mathsf P_G\|_{L^2(G)\to L^2(G)}\ge1.
\]
The reverse inequality was established above, so the error equals one. Since $n=\log_2|G|$, fewer than $\log_2|G|$ rewritings cannot achieve any operator error strictly below one for this family of groups. This proves the final claim.
\end{proof}

\section{Proof of \texorpdfstring{Corollary~\ref{thm:decoder-agnostic}}{Corollary 3}}
\label{app:proof-decoder-agnostic}

For completeness, the pathwise statement used in the main text is
\begin{equation}
    \|\mathsf A_S\widehat F_{x,\xi}-\mathsf P_G\widehat F_{x,\xi}\|_{L^2(G;\mathcal H)}
    \le \frac{8}{3}\sqrt{\frac{\log(|G|/\delta)}{|S|}}
    \|\widehat F_{x,\xi}-\mathsf P_G\widehat F_{x,\xi}\|_{L^2(G;\mathcal H)}.
    \label{eq:pathwise-decoder-law}
\end{equation}

\begin{proof}[Proof of Corollary~\ref{thm:decoder-agnostic}]
Fix the operator event in~\cref{thm:sparse-reynolds-full}. For any choices of $x$, $\theta$, $\tau$, $\mathcal D$, and $\mathcal H$, the decoded response $F_x^{\mathcal D}$ belongs to $L^2(G;\mathcal H)$. Applying~\cref{eq:hilbert-spectral-bound-full} to this function and using $q_x^{\mathcal D}=\mathsf P_GF_x^{\mathcal D}$ proves~\cref{eq:decoder-scaling-law}. The event depends only on $S$, so the conclusion is simultaneous over all of these choices.

For a randomized decoder, condition on any seed $\xi$ and all sampled traces. Its output is then a deterministic field $\widehat F_{x,\xi}\in L^2(G;\mathcal H)$. A second application of~\cref{eq:hilbert-spectral-bound-full} proves~\cref{eq:pathwise-decoder-law}. Because the operator event already holds for every function, the conclusion is simultaneous over all seeds and trace realizations and requires no union bound over $\Omega$.
\end{proof}

\section{Proof of \texorpdfstring{Corollary~\ref{cor:margin}}{Corollary 4}}
\label{app:proof-margin}

\begin{proof}[Proof of Corollary~\ref{cor:margin}]
If $\widehat c_S(h\cdot X)\ne y^\star(X)$, then either $c_G(X)\ne y^\star(X)$ or $\widehat c_S(h\cdot X)\ne c_G(X)$. Hence
\[
    R_S\le R_G+
    \mathbb P_{X,h}\bigl(\widehat c_S(h\cdot X)\ne c_G(X)\bigr).
\]
Split the second probability according to whether $\Delta_X\le t$. On $\{\Delta_X>t\}$, \cref{lem:margin-error} shows that a changed decision requires
\[
    \|(\mathsf A_SF_X)(h)-q_X\|_2^2\ge \frac{t^2}{2}.
\]
For each $X$ with $\Delta_X>t$, Markov's inequality over $h\sim\Unif(G)$ therefore gives
\begin{align*}
    \mathbb P_h\bigl(\widehat c_S(h\cdot X)\ne c_G(X),\ \Delta_X>t\mid X\bigr)
    &\le \frac{2}{t^2}
       \|\mathsf A_SF_X-\mathsf P_GF_X\|_{L^2(G;\mathbb R^C)}^2\\
    &\le \frac{128}{9t^2}
       \frac{\log(|G|/\delta)}{|S|}
       \|F_X-\mathsf P_GF_X\|_{L^2(G;\mathbb R^C)}^2,
\end{align*}
where the second line uses~\cref{eq:hilbert-spectral-bound-full}. Taking expectation over $X$ and adding $\mathbb P_X(\Delta_X\le t)$ proves~\cref{eq:risk-decomposition}.
\end{proof}

\input{sections/appendix_experiments}
\FloatBarrier

\input{sections/notation}

\end{document}

%% file: math_commands.tex
\usepackage{amsmath,amsfonts,bm}

\def\eqref#1{equation~\ref{#1}}
\def\1{\bm{1}}

\DeclareMathAlphabet{\mathsfit}{\encodingdefault}{\sfdefault}{m}{sl}
\SetMathAlphabet{\mathsfit}{bold}{\encodingdefault}{\sfdefault}{bx}{n}

%% file: figures/two-sources.tex
\definecolor{inkgray}{HTML}{2B2F36}
\definecolor{midgray}{HTML}{9CA1A8}
\definecolor{orbitfill}{HTML}{F6F7F8}
\definecolor{decodeblue}{HTML}{2A66B0}
\definecolor{rewriteorange}{HTML}{D8731F}

\newcommand{\orbit}{%
  \fill[orbitfill] (0,0) ellipse (2.15 and 0.78);
  \draw[inkgray, line width=0.8pt] (0,0) ellipse (2.15 and 0.78);
}
\newcommand{\faintpoints}{%
  \foreach \a in {20,55,90,125,160,200,240,300,340}{ \fill[midgray] ({2.15*cos(\a)},{0.78*sin(\a)}) circle (1.2pt); }
}
\newcommand{\fiber}[4]{%
  \draw[#4, line width=0.7pt] #1 -- ++(0,#2);
  \foreach \h in {#3}{ \fill[decodeblue] ($#1+(0,\h)$) circle (2.0pt); }
}
\newcommand{\panelcaption}[1]{\node[inkgray, font=\Large\itshape, align=center] at (0,-1.5) {#1};}

\begin{tikzpicture}[font=\Large, >={Latex[length=2.2mm,width=1.6mm]}]

\begin{scope}[shift={(-4.6,0)}]
  \coordinate (k) at (0,-1.0);
  \draw[rewriteorange, ->, line width=0.9pt] ({1.4*cos(200)},{-0.5+0.5*sin(200)}) arc[start angle=200, end angle=340, x radius=1.4, y radius=0.5];
  \draw[decodeblue, ->, line width=0.9pt] (k) -- (0,1.9);
  \fill[inkgray] (k) circle (1.8pt);
  \node[inkgray, anchor=south west] at (0.1,-0.98) {$x$};
  \node[decodeblue, anchor=south, align=center] at (0,1.95) {decoding\\[-2pt] reasoning traces $\omega$};
  \node[rewriteorange, anchor=north, align=center] at (0,-1.5) {rewriting\\[-2pt] along the orbit};
\end{scope}

\begin{scope}
  \orbit \faintpoints
  \coordinate (x0) at (0,-0.78);
  \fiber{(x0)}{2.55}{0.28,0.54,0.80,1.06,1.32,1.58,1.84,2.10,2.36}{decodeblue!55}
  \fill[rewriteorange] (x0) circle (2.4pt);
  \panelcaption{repeated decoding\\ one rep., many traces}
\end{scope}

\begin{scope}[shift={(5.9,0)}]
  \orbit
  \foreach \a in {5,20,35,50,65,80,95,110,125,140,155,170,185,200,215,230,245,260,275,290,305,320,335,350}{
     \coordinate (p) at ({2.15*cos(\a)},{0.78*sin(\a)});
     \fiber{(p)}{1.0}{0.32,0.66}{decodeblue!30}
     \fill[rewriteorange] (p) circle (1.9pt);
  }
  \panelcaption{full orbit\\ every rep.}
\end{scope}

\begin{scope}[shift={(11.8,0)}]
  \orbit \faintpoints
  \foreach \a in {205,250,290,335}{
     \coordinate (p) at ({2.15*cos(\a)},{0.78*sin(\a)});
     \fiber{(p)}{1.3}{0.36,0.72}{decodeblue!55}
     \fill[rewriteorange] (p) circle (2.4pt);
  }
  \panelcaption{sparse orbit\\ a few random reps.}
\end{scope}

\node[inkgray, anchor=north] at (3.9,-3.2)
  {\textcolor{rewriteorange}{\raisebox{0.4ex}{\scalebox{0.9}{$\bullet$}}}\; one rewriting $g\cdot x$ of the prompt
   \qquad
   \textcolor{decodeblue}{\raisebox{0.4ex}{\scalebox{0.75}{$\bullet$}}}\; one reasoning trace $\omega\sim p_\theta(\cdot\mid g\cdot x)$};

\end{tikzpicture}

%% file: figures/pipeline.tex
\definecolor{inkgray}{HTML}{2B2F36}
\definecolor{boxline}{HTML}{7A828C}
\definecolor{boxgray}{HTML}{F5F6F8}
\definecolor{decodeblue}{HTML}{2A66B0}
\definecolor{decodeblueL}{HTML}{EAF1FA}
\definecolor{rewriteorange}{HTML}{D8731F}
\definecolor{rewriteorangeL}{HTML}{FCEFE3}

\begin{tikzpicture}[
  font=\large,
  >={Latex[length=2.2mm,width=1.6mm]},
  flow/.style={draw=inkgray, line width=0.7pt, ->},
  thin/.style={draw=boxline, line width=0.55pt, ->},
  cell/.style={rounded corners=3pt, line width=0.55pt, minimum height=0.72cm, align=center, inner xsep=7pt},
  ocell/.style={cell, draw=rewriteorange!70, fill=rewriteorangeL},
  bcell/.style={cell, draw=decodeblue!70,   fill=decodeblueL},
  gcell/.style={cell, draw=boxline,          fill=boxgray},
  lane/.style={rounded corners=5pt, draw=boxline!55, line width=0.5pt, fill=none},
  head/.style={font=\large\itshape, text=inkgray, align=center},
  foot/.style={font=\normalsize\itshape, text=inkgray!80, align=center},
  dots/.style={font=\large, text=inkgray!70}
]

\def\rowA{0.95}\def\rowB{0}\def\rowC{-0.95}

\node[gcell, minimum height=1.0cm, minimum width=1.5cm] (x) at (0,0) {prompt\\[-2pt] $x$};

\node[ocell, minimum width=2.0cm] (g1) at (3.0,\rowA) {$g_1\cdot x$};
\node[dots] (gd) at (3.0,\rowB) {$\vdots$};
\node[ocell, minimum width=2.0cm] (gm) at (3.0,\rowC) {$g_m\cdot x$};
\node[lane, fit=(g1)(gm), inner sep=7pt] (G) {};
\node[head, above=0.15cm of G.north] {rewritten prompts};
\node[foot, below=0.15cm of G.south] {$y^\star(g\cdot x)=y^\star(x)$};

\node[bcell, minimum width=3.6cm] (p1) at (7.3,\rowA) {$\omega\sim p_{\theta,\tau}(\cdot\mid g_1\cdot x)$};
\node[dots] (pd) at (7.3,\rowB) {$\vdots$};
\node[bcell, minimum width=3.6cm] (pm) at (7.3,\rowC) {$\omega\sim p_{\theta,\tau}(\cdot\mid g_m\cdot x)$};
\node[lane, fit=(p1)(pm), inner sep=7pt] (P) {};
\node[head, above=0.15cm of P.north] {frozen model};
\node[foot, below=0.15cm of P.south] {one decoding fiber per rewriting};

\node[bcell, minimum width=2.1cm] (f1) at (11.5,\rowA) {$F_x^{\mathcal D}(g_1)$};
\node[dots] (fd) at (11.5,\rowB) {$\vdots$};
\node[bcell, minimum width=2.1cm] (fm) at (11.5,\rowC) {$F_x^{\mathcal D}(g_m)$};
\node[lane, fit=(f1)(fm), inner sep=7pt] (F) {};
\node[head, above=0.15cm of F.north] {decoded orbit field};
\node[foot, below=0.15cm of F.south] {$F_x^{\mathcal D}:G\to\mathcal H$};

\node[gcell, minimum width=3.0cm, minimum height=0.8cm] (full)   at (15.7,0.55)  {$\mathsf P_G F_x^{\mathcal D}=q_x^{\mathcal D}$};
\node[gcell, minimum width=3.0cm, minimum height=0.8cm, dash pattern=on 3pt off 2pt] (sparse) at (15.7,-0.55) {$\mathsf A_S F_x^{\mathcal D}\approx q_x^{\mathcal D}$};
\node[lane, fit=(full)(sparse), inner sep=7pt] (Q) {};
\node[head, above=0.15cm of Q.north] {symmetrization};
\node[foot, below=0.15cm of Q.south] {full: all $|G|$ rewritings\\[-2pt] sparse: $|S|$ random rewritings};

\draw[flow] (x.east) -- (G.west);
\draw[thin] (g1.east) -- (p1.west);
\draw[thin] (gm.east) -- (pm.west);
\draw[thin] (p1.east) -- node[above, font=\normalsize, text=inkgray!85] {$\mathcal D$} (f1.west);
\draw[thin] (pm.east) -- node[below, font=\normalsize, text=inkgray!85] {$\mathcal D$} (fm.west);
\draw[flow] (F.east) -- (Q.west);

\end{tikzpicture}

%% file: sections/experiments.tex
\section{Experiments}
\label{sec:experiments}

\rev{We test three predictions of the theory on frozen open-weight models that generate reasoning traces. Details and all results are in \cref{app:experiments}. The first prediction is that decoding at one representation cannot replace rewriting (\cref{thm:two-worlds}). With the same number of forward passes, resampling one prompt should saturate below the orbit average, which rewriting reaches. The second is that the number of rewritings needed grows at most logarithmically in $|G|$ (\cref{thm:sparse-reynolds,thm:decoder-agnostic}). The relative error of the sparse average should decay as $|S|^{-1/2}$ for every decoding rule and model, with a constant that grows at most like $\sqrt{\log|G|}$. The third comes from \cref{cor:margin}. Rewriting improves accuracy exactly on the instances where the model's answer depends on the representation \emph{and} the orbit-averaged decision is more accurate than a typical one, and it cannot help elsewhere.}

\paragraph{\rev{Setup.}}
\rev{Unless stated otherwise the model is Qwen2.5-7B-Instruct \citep{yang2024qwen25}, frozen, prompted to reason step by step and end with an answer line, which we parse. Each forward pass is one sampled trace at temperature $0.7$. We compare procedures at a matched budget of $B$ passes. \emph{Self consistency} spends all $B$ passes on the original prompt and takes a majority vote. \emph{Symmetrization} spends them on $m$ random rewritings with $B/m$ traces each and votes over all traces, so $m=1$ is self consistency and $m=B$ is one trace per rewriting. \emph{Greedy} is one deterministic pass. Rewritings are drawn once per instance and reused for every model (\cref{thm:decoder-agnostic}). Our main family of tasks is graph reasoning on Erd\H{o}s--R\'enyi graphs with $n$ nodes serialized as edge lists \citep{herbst2025lost,fatemi2024talk}. The group $G=S_n\times S_{|E|}$ acts by node relabeling and edge reordering, and we use seven invariant tasks (edge count, edge existence, node degree, connected components, connectivity, cycle existence, triangle count), with $100$ instances per task and size ($200$ in the cells of \cref{tab:exp-main}) and $n\in\{8,12,16,20,24\}$. We also use a non-graph task with two different groups, a multiple-choice benchmark with an equivariant answer, few-shot classification, and three further model families (Phi-4-mini, Llama-3.1-8B, Qwen2.5-14B-AWQ). Gains are paired differences in accuracy with $95\%$ bootstrap intervals over instances.}

\begin{figure}[t]
    \centering
    \includegraphics[width=\linewidth]{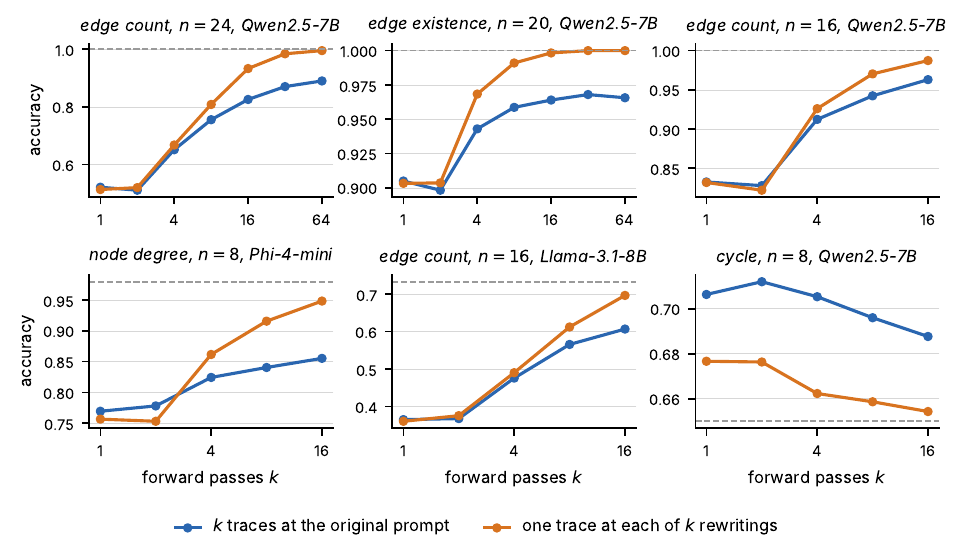}
    \caption{\rev{\textbf{Accuracy against the number of forward passes} (\cref{thm:two-worlds}). Accuracy against the number of forward passes $k$ when all $k$ traces are sampled at the original prompt and voted (blue), or one trace is sampled at each of $k$ random rewritings (orange), for six (task, size, model) cells. The dashed line is the vote over all traces at all rewritings. The last panel is the one task on which both aggregations hurt, discussed in the text.}}
    \label{fig:exp-curves}
\end{figure}

\paragraph{\rev{Rewriting versus decoding at matched compute.}}
\rev{\Cref{fig:exp-curves} shows the two budgets on the same instances. On edge counting with $n=24$, one greedy pass is right $39\%$ of the time. With $64$ traces at the original prompt the accuracy reaches $89\%$ and the curve flattens. With $64$ rewritings it reaches $99.5\%$, and $16$ rewritings already give $100\%$ against $83\%$ for $16$ traces, a gain of $+16.5$ points ($[+11.3,+22.2]$). The plateau is the content of \cref{thm:two-worlds}. More traces at one representation reveal nothing about the rest of the orbit. \Cref{tab:exp-main} reports the matched budget $B=16$ for four model families. Every family has significant gains, but on different tasks. Qwen2.5-7B gains on edge counting and edge existence, Phi-4-mini on all three tasks, Llama on edge counting, and the $14$B model on node degree. The rewriting set is the same for all of them, as \cref{thm:decoder-agnostic} allows, and which decision it improves is a property of the model (\cref{cor:margin}). Tokens per trace are the same at a rewriting as at the original prompt (ratio $0.999$, \cref{app:exp-setup}), so budgets in passes and in tokens coincide. The full table, with all seven tasks, all sizes, and the allocation grid over $(m,B/m)$, is in \cref{app:exp-main}. Across the $98$ (model, task, size) cells on these graphs, $24$ gains and $3$ losses are significant.}

\begin{table}[t]
    \centering
    \small
    \setlength{\tabcolsep}{3.5pt}
    \caption{\rev{Accuracy at $B=16$ forward passes on graphs with $n=16$ nodes ($n=24$ for Qwen2.5-7B). One greedy pass, self consistency with $16$ traces (SC), and $16$ rewritings with one trace each (Sym.), with the paired gain of Sym.\ over SC in accuracy points and its $95\%$ interval. The column ``flip'' is the fraction of instances whose greedy answer changes across the $16$ rewritings. Bold means that the interval excludes zero.}}
    \label{tab:exp-main}
    \vspace{4pt}
    \rev{\begin{tabular}{llcccrc}
        \toprule
        Model & Task & Greedy & SC & Sym. & Sym.\ $-$ SC & flip \\
        \midrule
        Qwen2.5-7B, $n{=}24$ & edge count     & 0.42 & 0.81 & 0.92 & $\mathbf{+10.8}$ $[+3.8,+18.4]$ & 0.94 \\
                             & edge existence & 0.88 & 0.96 & 1.00 & $\mathbf{+3.9}$ $[+0.9,+8.0]$ & 0.75 \\
                             & node degree    & 0.44 & 0.40 & 0.38 & $-2.2$ $[-8.0,+3.4]$ & 0.90 \\
        Llama-3.1-8B         & edge count     & 0.28 & 0.60 & 0.70 & $\mathbf{+10.0}$ $[+2.3,+18.3]$ & 1.00 \\
                             & edge existence & 0.72 & 0.81 & 0.86 & $+5.5$ $[-0.7,+12.3]$ & 0.57 \\
                             & node degree    & 0.68 & 0.88 & 0.90 & $+1.9$ $[-4.3,+8.4]$ & 0.87 \\
        Phi-4-mini           & edge count     & 0.71 & 0.89 & 0.99 & $\mathbf{+9.7}$ $[+5.5,+14.0]$ & 0.79 \\
                             & edge existence & 0.89 & 0.97 & 1.00 & $\mathbf{+2.4}$ $[+0.6,+4.5]$ & 0.67 \\
                             & node degree    & 0.68 & 0.75 & 0.82 & $\mathbf{+6.9}$ $[+2.2,+11.9]$ & 0.88 \\
        Qwen2.5-14B-AWQ      & edge count     & 0.63 & 0.71 & 0.69 & $-1.9$ $[-7.5,+3.8]$ & 0.52 \\
                             & edge existence & 0.98 & 0.99 & 1.00 & $+0.7$ $[+0.0,+2.0]$ & 0.17 \\
                             & node degree    & 0.83 & 0.85 & 0.94 & $\mathbf{+9.0}$ $[+2.1,+17.1]$ & 0.73 \\
        \bottomrule
    \end{tabular}}
\end{table}

\begin{figure}[t]
    \centering
    \includegraphics[width=\linewidth]{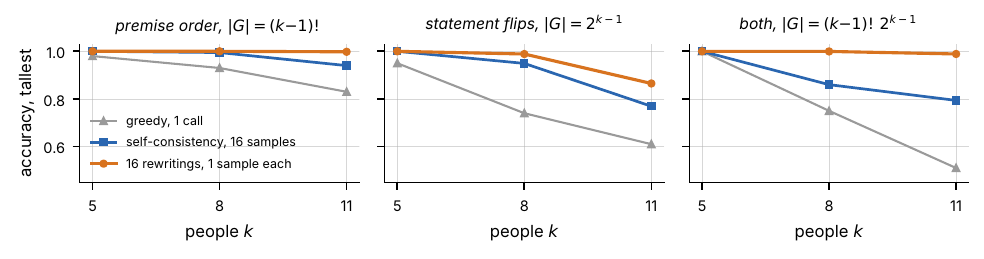}
    \caption{\rev{\textbf{A non-graph task under three groups.} ``Who is the tallest of $k$ people?'' given the $k-1$ adjacent comparisons, rewritten by permuting the premises ($S_{k-1}$), by flipping statements between ``taller than'' and ``shorter than'' ($\mathbb Z_2^{k-1}$), or both. Qwen2.5-7B, $100$ instances per $k$, budget $16$. The three ``position'' and ``comparison'' variants are in \cref{app:exp-premises}.}}
    \label{fig:exp-premises}
\end{figure}

\paragraph{\rev{Beyond graphs and beyond $S_n$.}}
\rev{\Cref{fig:exp-premises} repeats the comparison on a deduction task. Its rewritings form $S_{k-1}$, the hypercube $\mathbb Z_2^{k-1}$ on which the $\log|G|$ factor of \cref{thm:sparse-reynolds} is attained, or their product. With $k=11$ people and the product group, one greedy pass finds the tallest person $51\%$ of the time, $16$ traces $79\%$, and $16$ rewritings $99\%$ ($+19.5$ points, $[+11.9,+27.3]$). Across the nine (question, group) combinations and three values of $k$, $13$ of the $27$ cells are significant gains and none is a significant loss. For an equivariant answer we permute the four options of ARC-Challenge \citep{clark2018arc} ($G=S_4$, $500$ questions) and align the predicted letters by the inverse permutation before voting, as in \cref{sec:formulation}. The original ordering gives $91.8\%$ and the aligned vote over $16$ permutations gives $94.6\%$ ($+2.6$ over a matched vote at the original ordering, $[+0.8,+4.6]$). The same vote without alignment gives $24.4\%$, which is chance level. Reordering few-shot demonstrations on SST-2 \citep{lu2022fantastically}, finally, changes Qwen2.5-7B's answer on only $6$--$8\%$ of queries, and symmetrization changes accuracy by less than half a point in either direction, as the third prediction requires (\cref{app:exp-other}).}

\begin{figure}[t]
    \centering
    \includegraphics[width=\linewidth]{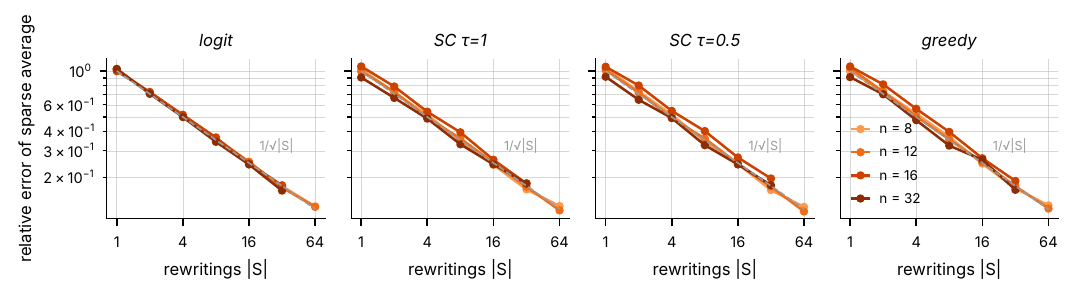}
    \caption{\rev{\textbf{Relative error of the sparse average on language-model fields} (\cref{thm:sparse-reynolds,thm:decoder-agnostic}). Relative error $\|\mathsf A_SF-\mathsf P_GF\|/\|F-\mathsf P_GF\|$ of the sparse average against $|S|$ for Qwen2.5-7B on graphs with $n\in\{8,12,16,32\}$ nodes, one shared $S$, four decoding rules applied to the same next-token fields, median over seven tasks. $\mathsf P_GF$ is estimated from $64$ independent rewritings and the estimation noise is subtracted (\cref{app:exp-rho}). The dashed line is $|S|^{-1/2}$.}}
    \label{fig:exp-rho}
\end{figure}

\paragraph{\rev{The number of rewritings needed as the group grows.}}
\rev{To measure the law itself we read next-token answer distributions, which give the decoded field $F_x^{\mathcal D}(g)$ exactly for every decoding rule. \Cref{fig:exp-rho} shows the relative error of $\mathsf A_SF$ for $n$ from $8$ to $32$, where $\log|G|$ grows from about $25$ to more than $300$ nats counting both factors of $G$. The rescaled error $\sqrt{|S|}\,\|\mathsf A_SF-\mathsf P_GF\|/\|F-\mathsf P_GF\|$ stays between $0.9$ and $1.1$ for every size, every task, and every decoding rule. The curves for Qwen2.5-1.5B on the same $S$ are indistinguishable from those of the $7$B model (\cref{app:exp-rho}). The logarithm in \cref{thm:sparse-reynolds} is a worst case over groups that node relabeling does not pay. On the accuracy side the same holds. On edge counting the fraction of instances whose answer changes under rewriting grows from $12\%$ at $n=8$ to $94\%$ at $n=24$, and the gain of $16$ rewritings over $16$ traces grows with it from $+1.0$ to $+10.8$ points, with the budget fixed at $16$ (\cref{app:exp-scaling}).}

\paragraph{\rev{When rewriting helps.}}
\rev{\Cref{cor:margin} says that rewriting can only recover competence the model shows on part of the orbit. Among the $98$ graph cells of the four models, $60$ are cells in which the greedy answer changes on more than half of the instances and the model is above $40\%$ accuracy. They contain $20$ significant gains and $1$ loss, with a mean gain of $+2.1$ points. The other $38$ cells contain $4$ gains and $2$ losses, with a mean of $0.0$ (\cref{app:exp-main}). Triangle counting is beyond every model at every representation. Counting components and connectivity are sensitive, but their orbit-averaged decision is no better than a typical one, so neither $16$ traces nor $16$ rewritings move them. The one systematic loss is cycle detection at $n=8$, where every model has a representation-invariant bias toward ``yes'' on acyclic graphs. A single greedy pass says ``yes'' on $48\%$ of them and the orbit on $62\%$. Averaging cannot remove an invariant bias, and when the wrong answer has probability above one half any vote, over traces or over rewritings, amplifies it. Both aggregations fall below greedy there (\cref{fig:exp-curves}, last panel). Conversely, no single serialization is a substitute for the orbit. On edge counting with $n=24$, only $6\%$ of the graphs are solved at all $16$ of their rewritings and a random rewriting solves $38\%$, while the vote over the $16$ solves $92\%$, nearly the $93\%$ an oracle choice of serialization would reach (\cref{app:exp-rewritings}). The same regime appears in a $0.8$M-parameter transformer trained from scratch on edge counting with a canonical serialization. It is perfect on its training format, $85\%$ accurate on a random rewriting, and $99.5\%$ after symmetrization over $16$ rewritings, while the same model trained with random rewritings is invariant and gains nothing (\cref{app:exp-scratch}). Finally, when the group is not available, rewritings can be sampled from a paraphrasing model instead. A cross-family paraphraser preserves the graph in $88\%$ of its rewordings at $n=12$ and $67\%$ at $n=24$. Over the surviving rewordings, symmetrization gains $+9.8$ points ($[+3.8,+15.9]$) on edge counting at $n=12$, but its orbit average stays below that of the exact group at $n=24$ ($87\%$ against $100\%$), and a same-model paraphraser, whose rewordings are near duplicates, gains nothing (\cref{app:exp-paraphrase}). Exact rewritings are cheaper and better when they exist, and paraphrases are a usable substitute whose error is the drift of the paraphraser.}

%% file: sections/appendix_experiments.tex
\section{Experimental Details and Additional Results}
\label{app:experiments}

\rev{This appendix gives the full setup of \cref{sec:experiments} and all results.}

\subsection{\rev{Setup}}
\label{app:exp-setup}

\paragraph{\rev{Models and generation.}}
\rev{We use Qwen2.5-7B-Instruct and Qwen2.5-1.5B-Instruct \citep{yang2024qwen25}, Phi-4-mini-instruct \citep{abdin2025phi4mini}, Llama-3.1-8B-Instruct \citep{grattafiori2024llama3}, and Qwen2.5-14B-Instruct in its $4$-bit AWQ quantization \citep{lin2024awq}, all frozen, with their chat templates, served with vLLM \citep{kwon2023vllm} in \texttt{bfloat16} (\texttt{float16} for the quantized model). Reasoning traces are sampled at temperature $0.7$ with a cap of $1024$ tokens ($1536$ for $n\ge 20$). A trace that reaches the cap without an answer line counts as wrong. Fewer than $1\%$ of traces are cut on every task except triangle counting ($4$--$13\%$) and Llama on triangle counting ($20$--$36\%$). Prompts ask the model to ``think step by step, keeping the reasoning brief'' and to end with \texttt{Answer: <value>}, which is parsed as a number, a yes/no, a name, or a letter. For every instance we draw $16$ rewritings from a seed that depends only on the task, the size, and the instance, so that every model sees the same rewriting set, and generate $4$ traces at each rewriting plus one greedy trace, with $16$ traces at the first rewriting so that self consistency with $16$ traces is available. Every allocation $(m,r)$ with $m\cdot r=B$ is obtained by subsampling this cache, averaged over $60$ random subsets. Tokens per trace are the same whether a trace is sampled at the original prompt or at a rewriting (ratio $0.999$, range $0.975$--$1.066$ over the $35$ Qwen2.5-7B cells), so a budget in forward passes is also a budget in generated tokens.}

\paragraph{\rev{Graphs and groups.}}
\rev{Graphs are drawn from $G(n,p)$ with $p$ uniform in $[0.15,0.6]$ for $n\le 8$, $[0.1,0.4]$ for $n\le 12$, and $[0.07,0.3]$ otherwise, rejecting graphs without edges. Yes/no tasks are balanced. A prompt lists the nodes and the edges as pairs with integer labels, followed by the question. A rewriting is a pair of a node relabeling and an edge-order permutation. The relabeling is applied first, each edge is written with its smaller label first and the list is sorted, and then the edge permutation is applied, so that $g\mapsto g\cdot x$ is an action of $S_n\times S_{|E|}$ on prompts. Questions that name nodes are relabeled with the graph. Instance pools are fixed per (task, size), so the $100$-instance and $200$-instance cells share their first $100$ instances. The stochastic block model uses two equal communities with within-community edge probability $2.5p$ and between-community probability $p/4$, and the planted-cycle graphs add a cycle through $3$ to $n/2$ random nodes to $G(n,p/2)$. The serialization variants use adjacency lists (``Node A is connected to B, D.'') or letter labels \citep{herbst2025lost}.}

\paragraph{\rev{Other tasks.}}
\rev{\emph{Deduction.} There are $k$ people with a hidden total order of heights. The $k-1$ adjacent comparisons are given as premises in a random order, half of them phrased as ``A is taller than B'' and half as ``B is shorter than A''. The three questions are who is the tallest, who is the $j$-th tallest ($2\le j\le k-1$), and whether one named person is taller than another non-adjacent one (balanced yes/no). Rewritings permute the premises ($S_{k-1}$), flip the phrasing of each premise ($\mathbb Z_2^{k-1}$), or both. \emph{ARC-Challenge} \citep{clark2018arc} uses $500$ random test questions with four options, all $24$ orderings of the options, and $2$ sampled traces and one greedy trace per ordering. The predicted letter is mapped back to the option it names before voting. \emph{SST-2} \citep{socher2013recursive} uses $200$ test sentences with $k\in\{4,8,16\}$ balanced training demonstrations, rewritten by permuting the demonstrations ($S_k$).}

\subsection{\rev{Rewriting versus decoding on graphs}}
\label{app:exp-main}

\rev{\Cref{tab:app-main-full} reports every (model, task, size) cell on Erd\H{o}s--R\'enyi graphs. \Cref{tab:app-allocation} gives the allocation grid at budget $16$ for Qwen2.5-7B. On edge counting and edge existence the all-self-consistency column ($m=1$) is the worst cell of every row, while on the tasks that are at chance or beyond the model the rows are flat. \Cref{tab:app-variants} repeats the comparison on structured graphs and other serializations. The gains replicate on both topologies. On edge counting at $n=24$ the gain is $+10.9$ on SBM graphs and $+6.8$ on planted-cycle graphs. Serialization changes which tasks are sensitive but not the rule. With adjacency lists, counting edges requires de-duplicating the lists and the model fails at it ($15\%$), while node degree becomes a lookup ($95\%$), and the significant gains move accordingly (node degree $+3.3$, connectivity $+1.4$).}

\rev{\Cref{cor:margin} predicts the structure of these tables. We call a cell sensitive if the greedy answer changes on more than half of its instances and competent if greedy accuracy exceeds $40\%$. Of the $98$ cells in \cref{tab:app-main-full}, the $60$ sensitive and competent ones contain $20$ significant gains and $1$ significant loss (mean $+2.1$ points), the $19$ sensitive but incompetent ones contain $1$ gain and no loss (mean $+0.1$), and the $19$ insensitive ones contain $3$ gains and $2$ losses (mean $-0.1$). The losses are cycle detection at $n=8$ for Qwen2.5-7B ($-3.4$), Phi-4-mini ($-6.9$), and Qwen2.5-14B ($-3.5$). On acyclic graphs all models answer ``yes'' more often than not under a random rewriting, a bias that does not depend on the representation. On those instances the greedy answer at the starting rewriting says ``yes'' $48\%$ of the time for Qwen2.5-7B, the greedy answers over the orbit $62\%$, and the sampled traces $63\%$. Any vote amplifies a majority, so both self consistency and symmetrization fall below a single greedy pass ($0.69$ and $0.65$ against $0.76$). This is the term $R_G$ of \cref{cor:margin} being larger than the error of a typical representation, and it is unrelated to order sensitivity.}

\input{figures/rebuttal/tab_main_full}

\begin{table}[!htb]
    \centering\small
    \caption{\rev{Allocation of a budget of $16$ forward passes between $m$ rewritings and $r=16/m$ traces per rewriting, Qwen2.5-7B, with a majority vote over all traces.}}
    \label{tab:app-allocation}
    \vspace{4pt}
    \input{figures/rebuttal/tab_allocation}
\end{table}

\input{figures/rebuttal/tab_variants}

\subsection{\rev{Scaling in the graph size}}
\label{app:exp-scaling}

\begin{figure}[h]
    \centering
    \includegraphics[width=0.8\linewidth]{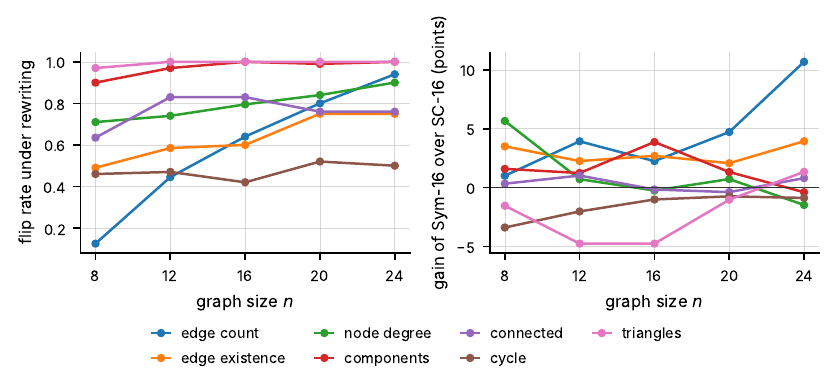}
    \caption{\rev{Order sensitivity and the gain of symmetrization against the graph size $n$, Qwen2.5-7B, budget $16$. The left panel shows the fraction of instances whose greedy answer changes across $16$ rewritings, and the right panel the paired gain of $16$ rewritings over $16$ traces.}}
    \label{fig:app-scaling}
\end{figure}

\rev{\Cref{fig:app-scaling} shows how sensitivity and gain depend on $n$ at a fixed budget. On edge counting the flip rate rises from $0.12$ at $n=8$ to $0.94$ at $n=24$ and the gain from $+1.0$ to $+10.8$ points. Edge existence is sensitive at every size and gains $+2$ to $+4$ points throughout. The cost of $16$ rewritings does not change with $n$, although $|G|$ grows from $8!\,|E|!$ to $24!\,|E|!$ across the figure. This is what the $\log|G|$ law of \cref{thm:sparse-reynolds} buys.}

\subsection{\rev{Per-rewriting accuracy}}
\label{app:exp-rewritings}

\begin{figure}[!htb]
    \centering
    \includegraphics[width=\linewidth]{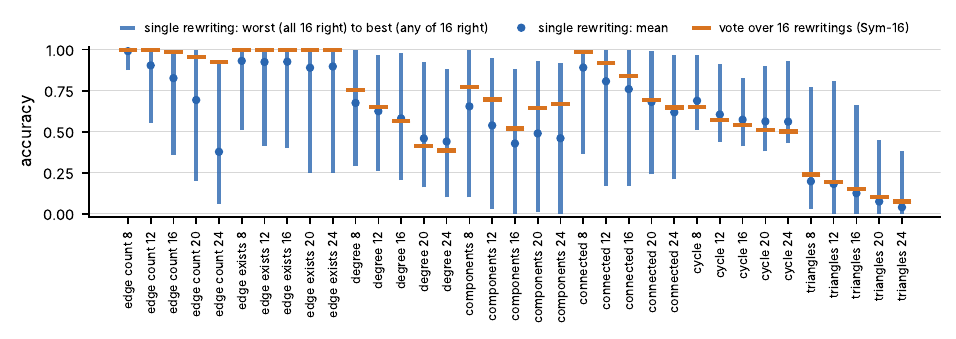}
    \caption{\rev{Single rewritings against their vote, Qwen2.5-7B, greedy decoding at each of the $16$ rewritings of every instance. The blue bar runs from the fraction of instances answered correctly at all $16$ rewritings (worst case, an adversary picks the serialization) to the fraction answered correctly at at least one (best case, an oracle picks it), and the dot is the accuracy of a random rewriting. The orange mark is Sym-16, the vote over one sampled trace at each of the $16$ rewritings.}}
    \label{fig:app-rewritings}
\end{figure}

\rev{\Cref{fig:app-rewritings} answers whether a well-chosen serialization could replace the average. For each instance we evaluate the greedy answer at each of its $16$ rewritings and record whether the model is right at all of them, at a random one, and at at least one. On edge counting at $n=24$ only $6\%$ of the graphs are solved at every rewriting and a random rewriting solves $38\%$, but some rewriting solves $93\%$, and the vote over the $16$ rewritings reaches $92\%$. The model has the answer on most graphs, just not at most representations, and the vote recovers nearly everything an oracle choice of serialization would. The same holds on edge existence ($25\%$, $89\%$, $100\%$ against a vote of $100\%$ at $n=20$). On node degree, component counting, connectivity, and cycle detection the oracle is also high ($88\%$ to $97\%$ at $n=24$) but the vote stays near the accuracy of a random rewriting. There, on a typical instance only a minority of rewritings are right, so the majority of the orbit is wrong and averaging converges, faithfully, to the wrong answer. This is the distinction \cref{cor:margin} draws between the accuracy of the symmetrized decision $c_G$ and the existence of some good representation. Symmetrization delivers the former, and only the former.}

\subsection{\rev{Deduction, multiple choice, and few-shot order}}
\label{app:exp-premises}

\begin{figure}[h]
    \centering
    \includegraphics[width=\linewidth]{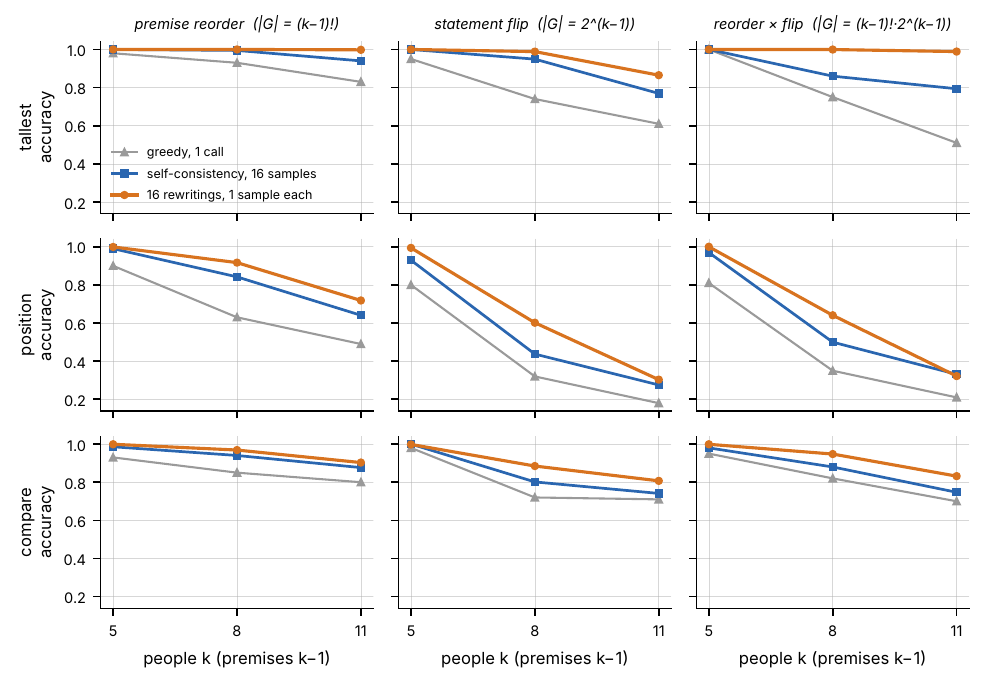}
    \caption{\rev{The deduction task under three groups, all three questions, Qwen2.5-7B, budget $16$. Each panel compares symmetrization with self consistency and greedy decoding on the same orbit.}}
    \label{fig:app-premises}
\end{figure}

\rev{\Cref{fig:app-premises,tab:app-premises} give the deduction task in full. Flipped statements are harder for the model than ``taller than'' statements, so the three groups have different baselines. Within each group symmetrization is significantly better than self consistency in $13$ of the $27$ cells and never significantly worse (the largest negative difference is $-0.9$ points, well inside its interval). The product group gives the largest gains, as it is both the largest orbit and the one whose rewritings differ most from one another.}

\input{figures/rebuttal/tab_premises}

\label{app:exp-other}
\begin{table}[!htb]
    \centering\small
    \caption{\rev{ARC-Challenge with permuted options, Qwen2.5-7B, $500$ questions, with $95\%$ bootstrap intervals over questions. ``Aligned'' maps each predicted letter back to the option it names before voting (\cref{sec:formulation}), and ``not aligned'' votes over the letters as written.}}
    \label{tab:app-arc}
    \vspace{4pt}
    \rev{\begin{tabular}{lc}
        \toprule
        Procedure & Accuracy \\
        \midrule
        Greedy, original ordering (1 call) & $0.918$ $[0.894,0.940]$ \\
        Vote over $3$ traces, original ordering & $0.922$ $[0.898,0.944]$ \\
        $4$ orderings, aligned, one trace each & $0.934$ $[0.914,0.956]$ \\
        $16$ orderings, aligned, one trace each & $0.946$ $[0.926,0.966]$ \\
        All $24$ orderings, aligned, all $72$ traces & $0.948$ $[0.928,0.966]$ \\
        $16$ orderings, \emph{not} aligned & $0.244$ $[0.206,0.282]$ \\
        \bottomrule
    \end{tabular}}
\end{table}

\rev{On ARC (\cref{tab:app-arc}) the per-ordering greedy accuracy ranges from $0.902$ to $0.932$ and $25.6\%$ of the questions change their answer under some ordering. Aligned symmetrization over $16$ orderings gains $+2.6$ points over the matched vote at the original ordering ($[+0.8,+4.6]$), and unaligned voting is at chance, as the equivariant setting of \cref{sec:formulation} requires. On SST-2 (\cref{tab:app-icl}) the model is at $96\%$ and only $6$--$8\%$ of queries are sensitive to the order of the demonstrations, so there is nothing to average and symmetrization changes accuracy by less than half a point.}

\input{figures/rebuttal/tab_icl}

\subsection{\rev{Rewritings from a paraphrasing model}}
\label{app:exp-paraphrase}

\begin{figure}[h]
    \centering
    \includegraphics[width=\linewidth]{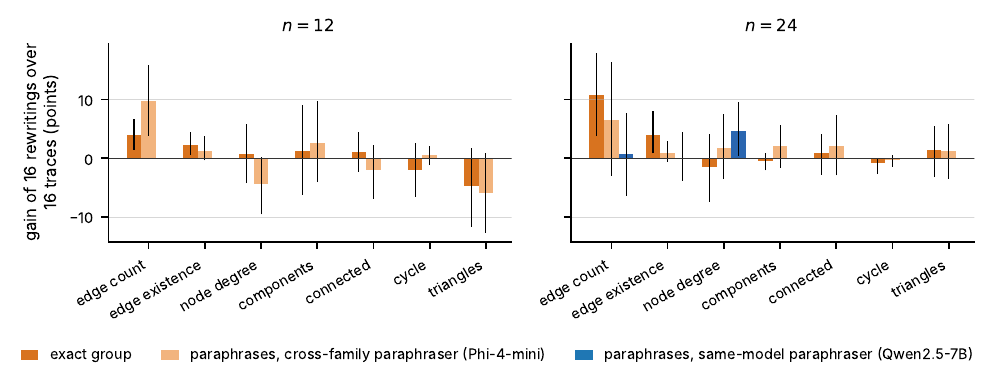}
    \caption{\rev{Gain of $16$ rewritings over $16$ traces on the same orbit, for the exact group and for the two paraphrasers, Qwen2.5-7B answering, with $95\%$ intervals over instances.}}
    \label{fig:app-paraphrase}
\end{figure}

\rev{When no group is available, rewritings can be sampled from a kernel $K(x,\cdot)$ given by a paraphrasing model. For rewritings drawn independently from $K$, $\mathsf A_S$ is an unbiased Monte Carlo estimate of the $K$-average at rate $|S|^{-1/2}$. What is lost relative to \cref{thm:sparse-reynolds} is that the target is meaningful only up to the drift $\varepsilon=\mathbb P_{x'\sim K(x)}[y^\star(x')\ne y^\star(x)]$, which enters \cref{cor:margin} as an additive term, and that there is no orbit over which to be uniform. We ask Phi-4-mini (cross-family) or Qwen2.5-7B itself (same model) for $24$ rewordings of the graph description, keep the question verbatim, and keep a rewording only if a second call that extracts its edge list returns exactly the true edge set. The rejection rate is our estimate of $\varepsilon$. The cross-family paraphraser keeps $21$ of $24$ rewordings at $n=12$ and $14$ of $24$ at $n=24$ ($\varepsilon\approx 12\%$ and $33\%$). The same-model paraphraser keeps $9$ of $24$, many of them near duplicates. \Cref{tab:app-paraphrase,fig:app-paraphrase} give the results. Gains over self consistency appear where the exact group gains ($+9.8$ on edge counting at $n=12$ and $+6.5$ at $n=24$). On the same instances, however, the paraphrase orbit average is below the group orbit average ($0.87$ against $1.00$ on edge counting at $n=24$, $0.79$ against $0.94$ on connectivity at $n=12$), both because of drift and because reworded descriptions are harder for the model than the canonical one. The same-model paraphrases gain nothing on edge counting ($+0.7$). Diversity of the rewriting distribution is what matters, and the group provides it exactly.}

\input{figures/rebuttal/tab_paraphrase}

\subsection{\rev{A model trained from scratch}}
\label{app:exp-scratch}

\begin{figure}[h]
    \centering
    \includegraphics[width=0.6\linewidth]{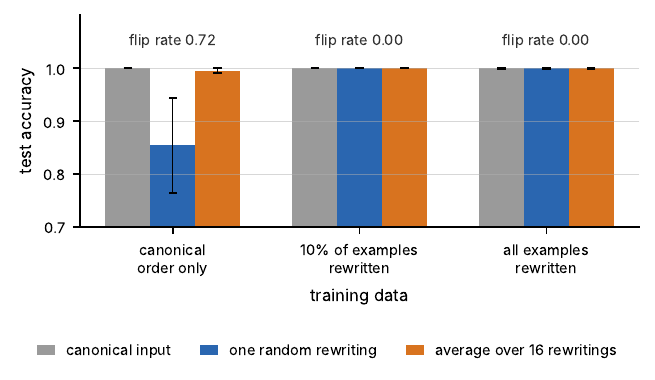}
    \caption{\rev{A $0.8$M-parameter transformer encoder trained from scratch to count the edges of graphs with $n=10$ nodes, under three training regimes, evaluated on $2{,}000$ held-out graphs, with mean and standard deviation over two seeds. ``flip'' is the fraction of test graphs whose prediction changes across $16$ rewritings.}}
    \label{fig:app-scratch}
\end{figure}

\rev{To isolate the mechanism from any property of pretrained models, we train a four-layer transformer encoder ($d=128$, $0.8$M parameters) on edge counting for graphs with $n=10$ nodes, $6{,}000$ steps of batch $256$ on freshly sampled graphs, under three regimes. The first uses the canonical serialization only, the second rewrites $10\%$ of the training examples at random, and the third rewrites every training example at random. \Cref{fig:app-scratch} shows the three cases of \cref{cor:margin} in one model. The canonical-only model is perfect on its training format and $85\%$ accurate on a random rewriting, with $72\%$ of test graphs changing their prediction across rewritings. Averaging its class probabilities over $16$ rewritings gives $99.5\%$ and over $64$ gives $100\%$, and its relative error follows $|S|^{-1/2}$ to two decimals ($1.00$, $0.71$, $0.50$, $0.35$ for $|S|=1,2,4,8$). Both augmented models are invariant (flip rate $0.000$) and gain nothing. On the node-degree task training is bimodal at this budget (some seeds converge to $100\%$ and others stall near $55\%$, in every regime), so we report edge counting only. The seeds that converge on node degree show the same pattern.}

\subsection{\rev{Relative error of language-model fields at large $n$}}
\label{app:exp-rho}

\begin{figure}[h]
    \centering
    \includegraphics[width=\linewidth]{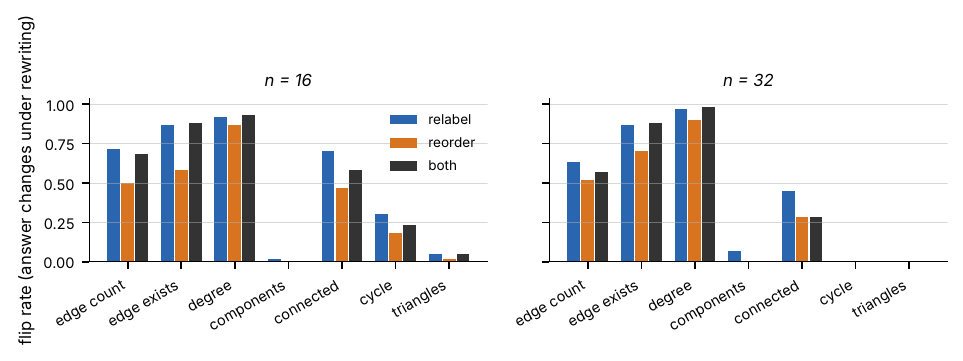}
    \caption{\rev{Fraction of graphs whose single-token answer changes under node relabeling only, edge reordering only, and both, Qwen2.5-7B, $60$ graphs per task.}}
    \label{fig:app-ablation}
\end{figure}

\rev{For \cref{fig:exp-rho} we prompt the model to answer with a single token and read its next-token distribution over the answer classes, which gives $F_x^{\mathcal D}(g)$ exactly for every decoding rule (expected logits, self consistency probabilities at $\tau\in\{0.5,1\}$, greedy). For $n\ge 8$ the orbit cannot be enumerated, so we draw one multiset $S$ of $32$ rewritings and $2$ starting rewritings per graph, and estimate $\mathsf P_GF$ from $64$ independent rewritings. The estimate adds a known variance $\|F-\mathsf P_GF\|^2/64$ to the numerator, which we subtract. The ratio is pooled over the $60$ graphs of each task and the median over tasks is plotted. The rescaled error $\sqrt{|S|}\,\|\mathsf A_SF-\mathsf P_GF\|/\|F-\mathsf P_GF\|$ lies in $[0.9,1.1]$ for all $n\in\{8,12,16,32\}$, decoders, and tasks whose fields are not nearly constant. For Qwen2.5-1.5B on the same $S$ it lies in $[0.96,1.09]$. \Cref{fig:app-ablation} separates the two factors of the group. Relabeling nodes moves the answer more than reordering edges on every task ($87\%$ against $70\%$ of instances on edge existence at $n=32$), and reordering alone is substantial on the edge-level tasks.}

\subsection{Computing the uniform constant}
\label{app:exp-opnorm}

\begin{figure}[h]
    \centering
    \includegraphics[width=0.62\linewidth]{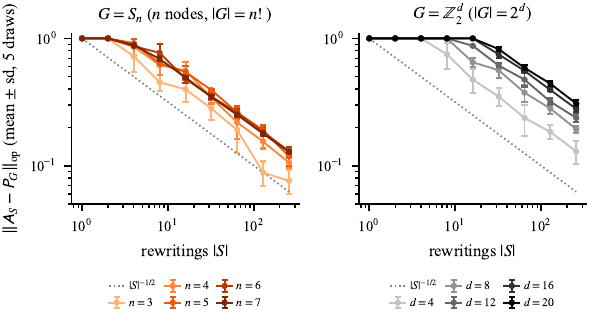}
    \caption{Exact uniform constant $\|\mathsf A_S-\mathsf P_G\|_{\mathrm{op}}$ of \cref{thm:sparse-reynolds} against $|S|$ for $G=S_n$ (graphs on $n=3,\dots,7$ nodes, $|G|=n!$) and $G=\mathbb Z_2^d$ ($|G|=2^d$, $d=4,\dots,20$); mean and standard deviation over five draws of $S$. For $\mathbb Z_2^d$ the constant stays at one until $|S|$ is of order $d$, and larger groups need proportionally more rewritings; for $S_n$ the curves nearly coincide.}
    \label{fig:app-opnorm}
\end{figure}

\begin{figure}[h]
    \centering
    \includegraphics[width=0.9\linewidth]{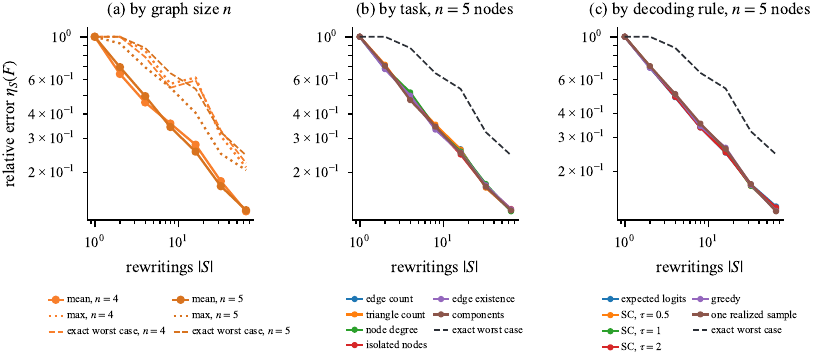}
    \caption{Relative error $\eta_S(F)$ for Qwen2.5-1.5B fields with one shared $S$ per size and draw. (a) Mean and maximum over fields, and the exact worst case of \cref{thm:sparse-reynolds}, for graphs with $n=4$ nodes ($|G|=24$) and $n=5$ nodes ($|G|=120$). (b) Mean by task and (c) by decoding rule at $n=5$. The curves are nearly indistinguishable, as \cref{thm:decoder-agnostic} predicts.}
    \label{fig:app-opnorm-lm}
\end{figure}

For a multiset $S$, the operator $\mathsf A_S-\mathsf P_G$ acts on $L^2(G)$ with uniform measure, and $\mathsf A_S-\mathsf P_G=(\mathsf A_S-\mathsf P_G)(I-\mathsf P_G)$ since $\mathsf A_S\mathsf P_G=\mathsf P_G\mathsf A_S=\mathsf P_G$. Hence
\[
    \sup_{F\notin\mathrm{im}\,\mathsf P_G}\frac{\|\mathsf A_SF-\mathsf P_GF\|}{\|F-\mathsf P_GF\|}
    =\|\mathsf A_S-\mathsf P_G\|_{\mathrm{op}}
\]
for scalar fields, and the same value holds for Hilbert-valued fields by applying the scalar bound coordinatewise in an orthonormal basis. For $G=S_n$ we represent functions on $G$ as vectors indexed by the lexicographic rank of permutations, implement $\mathsf A_S$ and its adjoint by index maps $h\mapsto s_ih$, and compute the largest singular value with ARPACK \citep{lehoucq1998arpack} through SciPy \citep{virtanen2020scipy}. For $n\le 5$ we checked every draw against a dense singular value decomposition; all $70$ draws agree to $10^{-6}$. For $G=\mathbb Z_2^d$ every irreducible representation is a character \citep{diaconis1988group} $\chi_T(s)=(-1)^{\langle T,s\rangle}$, so
$\|\mathsf A_S-\mathsf P_G\|_{\mathrm{op}}=\max_{T\ne 0}\bigl|\tfrac1{|S|}\sum_i\chi_T(s_i)\bigr|$, which we read off the Walsh--Hadamard transform of the empirical distribution of $S$ in $O(d\,2^d)$ time. \Cref{fig:app-opnorm} shows the full curves with their spread over five draws. \rev{This constant is the worst case over all fields, and \cref{fig:app-opnorm-lm} shows that on the exactly enumerated orbits of graphs with $4$ and $5$ nodes, no language-model field exceeds it for $|S|>1$, for Qwen2.5-1.5B with six decoding rules.}

%% file: figures/rebuttal/tab_main_full.tex
\begingroup\small\setlength{\tabcolsep}{4pt}
\begin{longtable}{llrccccrc}
\caption{Accuracy at a budget of $16$ forward passes on Erd\H{o}s--R\'enyi graphs for four model families: one greedy pass at a random rewriting, self consistency with $16$ traces at a random rewriting (SC-16), and one trace at each of $16$ rewritings (Sym.-16); paired gain (Sym.-16 minus SC-16) in points with $95\%$ bootstrap interval over the $N$ instances (bold: interval excludes zero); ``flip'' is the fraction of instances whose greedy answer changes across the $16$ rewritings.}\label{tab:app-main-full}\\
\toprule
Model & Task & $n$ & $N$ & Greedy & SC-16 & Sym.-16 & Gain & flip \\ \midrule \endfirsthead
\toprule
Model & Task & $n$ & $N$ & Greedy & SC-16 & Sym.-16 & Gain & flip \\ \midrule \endhead
\bottomrule \endfoot
Qwen2.5-7B & edge count & 8 & 200 & 0.99 & 0.99 & 1.00 & $+1.0$ $[+0.0,+2.5]$ & 0.12 \\
 & edge count & 12 & 200 & 0.92 & 0.96 & 1.00 & $\mathbf{+3.9}$ $[+1.4,+6.7]$ & 0.45 \\
 & edge count & 16 & 200 & 0.81 & 0.96 & 0.98 & $+2.2$ $[-0.2,+5.0]$ & 0.64 \\
 & edge count & 20 & 200 & 0.70 & 0.91 & 0.96 & $\mathbf{+4.7}$ $[+1.2,+8.7]$ & 0.80 \\
 & edge count & 24 & 100 & 0.42 & 0.82 & 0.92 & $\mathbf{+10.7}$ $[+3.8,+18.0]$ & 0.94 \\
 & edge existence & 8 & 200 & 0.94 & 0.96 & 1.00 & $\mathbf{+3.5}$ $[+1.2,+6.0]$ & 0.49 \\
 & edge existence & 12 & 200 & 0.92 & 0.98 & 1.00 & $\mathbf{+2.2}$ $[+0.5,+4.5]$ & 0.58 \\
 & edge existence & 16 & 200 & 0.92 & 0.97 & 1.00 & $\mathbf{+2.7}$ $[+0.8,+4.9]$ & 0.60 \\
 & edge existence & 20 & 200 & 0.85 & 0.98 & 1.00 & $\mathbf{+2.1}$ $[+0.3,+4.1]$ & 0.75 \\
 & edge existence & 24 & 100 & 0.88 & 0.96 & 1.00 & $\mathbf{+3.9}$ $[+0.8,+8.0]$ & 0.75 \\
 & node degree & 8 & 200 & 0.68 & 0.70 & 0.75 & $\mathbf{+5.7}$ $[+0.4,+10.9]$ & 0.71 \\
 & node degree & 12 & 200 & 0.61 & 0.64 & 0.65 & $+0.7$ $[-4.3,+5.9]$ & 0.74 \\
 & node degree & 16 & 200 & 0.59 & 0.57 & 0.56 & $-0.3$ $[-5.9,+5.3]$ & 0.80 \\
 & node degree & 20 & 200 & 0.42 & 0.40 & 0.41 & $+0.7$ $[-4.4,+5.6]$ & 0.84 \\
 & node degree & 24 & 100 & 0.44 & 0.40 & 0.39 & $-1.5$ $[-7.4,+4.1]$ & 0.90 \\
 & components & 8 & 100 & 0.67 & 0.76 & 0.77 & $+1.6$ $[-4.9,+8.0]$ & 0.90 \\
 & components & 12 & 100 & 0.52 & 0.68 & 0.70 & $+1.2$ $[-6.2,+9.1]$ & 0.97 \\
 & components & 16 & 100 & 0.39 & 0.48 & 0.52 & $+3.9$ $[-1.6,+9.1]$ & 1.00 \\
 & components & 20 & 100 & 0.47 & 0.63 & 0.65 & $+1.3$ $[-1.2,+3.3]$ & 0.99 \\
 & components & 24 & 100 & 0.44 & 0.67 & 0.67 & $-0.4$ $[-2.1,+1.0]$ & 1.00 \\
 & connected & 8 & 200 & 0.89 & 0.98 & 0.99 & $+0.3$ $[-1.1,+2.0]$ & 0.64 \\
 & connected & 12 & 200 & 0.84 & 0.91 & 0.92 & $+1.0$ $[-2.3,+4.5]$ & 0.83 \\
 & connected & 16 & 200 & 0.80 & 0.84 & 0.84 & $-0.2$ $[-3.7,+3.9]$ & 0.83 \\
 & connected & 20 & 200 & 0.69 & 0.70 & 0.69 & $-0.4$ $[-3.2,+2.4]$ & 0.76 \\
 & connected & 24 & 100 & 0.61 & 0.64 & 0.65 & $+0.8$ $[-2.8,+4.1]$ & 0.76 \\
 & cycle & 8 & 100 & 0.76 & 0.69 & 0.65 & $\mathbf{-3.4}$ $[-6.8,-0.6]$ & 0.46 \\
 & cycle & 12 & 100 & 0.59 & 0.59 & 0.57 & $-2.0$ $[-6.6,+2.6]$ & 0.47 \\
 & cycle & 16 & 100 & 0.59 & 0.55 & 0.54 & $-1.0$ $[-4.5,+1.4]$ & 0.42 \\
 & cycle & 20 & 100 & 0.58 & 0.52 & 0.51 & $-0.8$ $[-3.1,+0.7]$ & 0.52 \\
 & cycle & 24 & 100 & 0.56 & 0.51 & 0.50 & $-0.9$ $[-2.7,+0.1]$ & 0.50 \\
 & triangles & 8 & 100 & 0.22 & 0.25 & 0.24 & $-1.5$ $[-7.9,+4.2]$ & 0.97 \\
 & triangles & 12 & 100 & 0.23 & 0.24 & 0.20 & $-4.8$ $[-11.7,+1.7]$ & 1.00 \\
 & triangles & 16 & 100 & 0.10 & 0.20 & 0.15 & $-4.8$ $[-10.8,+1.4]$ & 1.00 \\
 & triangles & 20 & 100 & 0.11 & 0.11 & 0.10 & $-1.0$ $[-7.0,+4.0]$ & 1.00 \\
 & triangles & 24 & 100 & 0.06 & 0.06 & 0.07 & $+1.3$ $[-3.2,+5.4]$ & 1.00 \\
\midrule
Phi-4-mini & edge count & 8 & 200 & 0.98 & 0.99 & 1.00 & $+0.5$ $[+0.0,+1.5]$ & 0.27 \\
 & edge count & 12 & 200 & 0.87 & 0.95 & 0.99 & $\mathbf{+4.8}$ $[+1.9,+7.9]$ & 0.68 \\
 & edge count & 16 & 200 & 0.71 & 0.89 & 0.99 & $\mathbf{+9.7}$ $[+5.5,+14.0]$ & 0.79 \\
 & edge existence & 8 & 200 & 0.94 & 0.98 & 1.00 & $\mathbf{+1.7}$ $[+0.1,+3.5]$ & 0.51 \\
 & edge existence & 12 & 200 & 0.93 & 0.98 & 1.00 & $\mathbf{+1.8}$ $[+0.3,+3.8]$ & 0.58 \\
 & edge existence & 16 & 200 & 0.89 & 0.97 & 1.00 & $\mathbf{+2.4}$ $[+0.6,+4.5]$ & 0.67 \\
 & node degree & 8 & 200 & 0.73 & 0.85 & 0.95 & $\mathbf{+9.3}$ $[+4.6,+14.1]$ & 0.72 \\
 & node degree & 12 & 200 & 0.66 & 0.76 & 0.86 & $\mathbf{+9.9}$ $[+4.6,+15.1]$ & 0.77 \\
 & node degree & 16 & 200 & 0.68 & 0.75 & 0.82 & $\mathbf{+6.9}$ $[+2.2,+11.9]$ & 0.88 \\
 & components & 8 & 100 & 0.69 & 0.76 & 0.76 & $+0.2$ $[-5.7,+6.3]$ & 0.73 \\
 & components & 12 & 100 & 0.51 & 0.62 & 0.61 & $-1.1$ $[-6.0,+3.2]$ & 0.87 \\
 & components & 16 & 60 & 0.43 & 0.44 & 0.44 & $-0.0$ $[-2.8,+2.5]$ & 0.93 \\
 & connected & 8 & 100 & 0.81 & 0.76 & 0.79 & $+2.8$ $[-2.6,+8.2]$ & 0.58 \\
 & connected & 12 & 100 & 0.62 & 0.74 & 0.75 & $+1.0$ $[-5.5,+7.2]$ & 0.79 \\
 & connected & 16 & 60 & 0.55 & 0.64 & 0.62 & $-1.7$ $[-8.2,+4.3]$ & 0.80 \\
 & cycle & 8 & 100 & 0.68 & 0.83 & 0.76 & $\mathbf{-6.9}$ $[-12.6,-1.5]$ & 0.88 \\
 & cycle & 12 & 100 & 0.57 & 0.63 & 0.67 & $+4.1$ $[-0.9,+9.1]$ & 0.98 \\
 & cycle & 16 & 60 & 0.62 & 0.63 & 0.62 & $-0.6$ $[-7.8,+6.2]$ & 0.98 \\
 & triangles & 8 & 100 & 0.34 & 0.41 & 0.44 & $+2.9$ $[-2.7,+8.9]$ & 1.00 \\
 & triangles & 12 & 100 & 0.22 & 0.30 & 0.31 & $+0.6$ $[-5.9,+7.4]$ & 1.00 \\
 & triangles & 16 & 60 & 0.10 & 0.23 & 0.21 & $-2.1$ $[-8.2,+3.4]$ & 1.00 \\
\midrule
Llama-3.1-8B & edge count & 8 & 100 & 0.74 & 0.91 & 0.96 & $\mathbf{+5.7}$ $[+2.0,+10.4]$ & 0.98 \\
 & edge count & 12 & 100 & 0.36 & 0.81 & 0.79 & $-2.0$ $[-8.3,+4.6]$ & 0.99 \\
 & edge count & 16 & 60 & 0.28 & 0.61 & 0.70 & $\mathbf{+8.9}$ $[+0.9,+16.9]$ & 1.00 \\
 & edge existence & 8 & 100 & 0.84 & 0.94 & 0.98 & $\mathbf{+4.0}$ $[+0.1,+8.6]$ & 0.45 \\
 & edge existence & 12 & 100 & 0.84 & 0.91 & 0.91 & $-0.2$ $[-5.3,+4.6]$ & 0.50 \\
 & edge existence & 16 & 60 & 0.72 & 0.81 & 0.86 & $+5.3$ $[-1.0,+12.2]$ & 0.57 \\
 & node degree & 8 & 100 & 0.79 & 0.89 & 0.97 & $\mathbf{+7.8}$ $[+2.8,+13.4]$ & 0.83 \\
 & node degree & 12 & 100 & 0.74 & 0.85 & 0.90 & $+4.8$ $[-1.1,+11.2]$ & 0.90 \\
 & node degree & 16 & 60 & 0.68 & 0.88 & 0.90 & $+2.0$ $[-4.1,+8.8]$ & 0.87 \\
 & components & 8 & 100 & 0.52 & 0.67 & 0.64 & $-2.3$ $[-9.9,+5.1]$ & 1.00 \\
 & components & 12 & 100 & 0.37 & 0.51 & 0.51 & $+0.3$ $[-8.7,+9.3]$ & 1.00 \\
 & components & 16 & 60 & 0.28 & 0.41 & 0.38 & $-3.3$ $[-13.5,+6.4]$ & 1.00 \\
 & connected & 8 & 100 & 0.72 & 0.74 & 0.71 & $-2.3$ $[-6.7,+1.7]$ & 0.77 \\
 & connected & 12 & 100 & 0.54 & 0.58 & 0.57 & $-0.8$ $[-4.1,+2.3]$ & 0.91 \\
 & connected & 16 & 60 & 0.50 & 0.53 & 0.52 & $-1.3$ $[-6.3,+2.3]$ & 0.90 \\
 & cycle & 8 & 100 & 0.56 & 0.52 & 0.53 & $+0.9$ $[-1.0,+3.0]$ & 0.91 \\
 & cycle & 12 & 100 & 0.45 & 0.50 & 0.50 & $+0.0$ $[+0.0,+0.1]$ & 0.92 \\
 & cycle & 16 & 60 & 0.43 & 0.50 & 0.50 & $+0.0$ $[+0.0,+0.0]$ & 0.97 \\
 & triangles & 8 & 100 & 0.02 & 0.05 & 0.06 & $+0.7$ $[-2.7,+3.9]$ & 1.00 \\
 & triangles & 12 & 100 & 0.02 & 0.03 & 0.02 & $-0.2$ $[-3.5,+2.4]$ & 1.00 \\
 & triangles & 16 & 60 & 0.02 & 0.01 & 0.01 & $-0.1$ $[-0.9,+0.6]$ & 1.00 \\
\midrule
Qwen2.5-14B-AWQ & edge count & 8 & 100 & 1.00 & 0.99 & 1.00 & $+0.5$ $[+0.0,+1.7]$ & 0.03 \\
 & edge count & 12 & 100 & 0.79 & 0.80 & 0.78 & $-1.5$ $[-4.7,+1.0]$ & 0.36 \\
 & edge count & 16 & 60 & 0.63 & 0.71 & 0.70 & $-1.2$ $[-6.7,+4.2]$ & 0.52 \\
 & edge existence & 8 & 100 & 1.00 & 1.00 & 1.00 & $+0.0$ $[+0.0,+0.0]$ & 0.04 \\
 & edge existence & 12 & 100 & 0.98 & 1.00 & 1.00 & $+0.0$ $[+0.0,+0.0]$ & 0.14 \\
 & edge existence & 16 & 60 & 0.98 & 0.99 & 1.00 & $+0.7$ $[+0.0,+2.2]$ & 0.17 \\
 & node degree & 8 & 100 & 0.90 & 0.90 & 1.00 & $\mathbf{+10.1}$ $[+4.3,+16.2]$ & 0.55 \\
 & node degree & 12 & 100 & 0.78 & 0.82 & 0.94 & $\mathbf{+12.1}$ $[+5.9,+19.2]$ & 0.65 \\
 & node degree & 16 & 60 & 0.83 & 0.85 & 0.94 & $\mathbf{+8.9}$ $[+2.0,+16.5]$ & 0.73 \\
 & components & 8 & 100 & 0.80 & 0.86 & 0.87 & $+1.5$ $[-4.0,+7.0]$ & 0.58 \\
 & components & 12 & 100 & 0.68 & 0.71 & 0.70 & $-0.9$ $[-6.1,+4.0]$ & 0.87 \\
 & components & 16 & 60 & 0.53 & 0.54 & 0.50 & $-4.4$ $[-11.9,+2.2]$ & 0.90 \\
 & connected & 8 & 100 & 0.90 & 0.93 & 0.91 & $-2.8$ $[-7.4,+1.6]$ & 0.38 \\
 & connected & 12 & 100 & 0.72 & 0.82 & 0.80 & $-1.7$ $[-5.7,+1.9]$ & 0.53 \\
 & connected & 16 & 60 & 0.72 & 0.73 & 0.71 & $-2.0$ $[-8.5,+4.1]$ & 0.58 \\
 & cycle & 8 & 100 & 0.70 & 0.70 & 0.66 & $\mathbf{-3.5}$ $[-7.0,-0.6]$ & 0.30 \\
 & cycle & 12 & 100 & 0.62 & 0.63 & 0.59 & $-3.5$ $[-7.8,+0.1]$ & 0.40 \\
 & cycle & 16 & 60 & 0.57 & 0.55 & 0.57 & $+2.6$ $[-0.2,+6.6]$ & 0.32 \\
 & triangles & 8 & 100 & 0.28 & 0.31 & 0.36 & $+5.5$ $[-0.2,+11.6]$ & 0.95 \\
 & triangles & 12 & 100 & 0.17 & 0.21 & 0.23 & $+2.0$ $[-5.1,+8.5]$ & 0.99 \\
 & triangles & 16 & 60 & 0.12 & 0.26 & 0.22 & $-4.3$ $[-14.1,+5.0]$ & 1.00 \\
\end{longtable}
\endgroup

%% file: figures/rebuttal/tab_allocation.tex
\begin{tabular}{llcccc}
\toprule
Task & $n$ & $m{=}1$ ($r{=}16$) & $m{=}4$ ($r{=}4$) & $m{=}8$ ($r{=}2$) & $m{=}16$ ($r{=}1$) \\ \midrule
edge count & 8 & 0.990 & 1.000 & 1.000 & 1.000 \\
 & 12 & 0.958 & 0.995 & 0.998 & 0.999 \\
 & 16 & 0.963 & 0.984 & 0.985 & 0.988 \\
 & 20 & 0.910 & 0.948 & 0.959 & 0.961 \\
 & 24 & 0.817 & 0.912 & 0.927 & 0.930 \\
edge existence & 8 & 0.963 & 0.999 & 1.000 & 1.000 \\
 & 12 & 0.977 & 0.999 & 1.000 & 1.000 \\
 & 16 & 0.973 & 1.000 & 1.000 & 1.000 \\
 & 20 & 0.977 & 0.991 & 0.993 & 0.995 \\
 & 24 & 0.960 & 1.000 & 0.998 & 1.000 \\
node degree & 8 & 0.695 & 0.733 & 0.750 & 0.760 \\
 & 12 & 0.636 & 0.637 & 0.636 & 0.654 \\
 & 16 & 0.570 & 0.566 & 0.560 & 0.557 \\
 & 20 & 0.403 & 0.430 & 0.413 & 0.403 \\
 & 24 & 0.398 & 0.373 & 0.367 & 0.385 \\
components & 8 & 0.765 & 0.768 & 0.762 & 0.785 \\
 & 12 & 0.683 & 0.683 & 0.673 & 0.707 \\
 & 16 & 0.485 & 0.518 & 0.518 & 0.517 \\
 & 20 & 0.630 & 0.642 & 0.647 & 0.645 \\
 & 24 & 0.672 & 0.672 & 0.668 & 0.673 \\
connected & 8 & 0.983 & 0.976 & 0.988 & 0.993 \\
 & 12 & 0.907 & 0.925 & 0.923 & 0.917 \\
 & 16 & 0.838 & 0.828 & 0.833 & 0.834 \\
 & 20 & 0.696 & 0.698 & 0.689 & 0.699 \\
 & 24 & 0.642 & 0.667 & 0.648 & 0.655 \\
cycle & 8 & 0.683 & 0.660 & 0.655 & 0.655 \\
 & 12 & 0.592 & 0.563 & 0.577 & 0.567 \\
 & 16 & 0.550 & 0.547 & 0.538 & 0.535 \\
 & 20 & 0.512 & 0.505 & 0.508 & 0.508 \\
 & 24 & 0.510 & 0.502 & 0.502 & 0.500 \\
triangles & 8 & 0.253 & 0.230 & 0.248 & 0.225 \\
 & 12 & 0.245 & 0.195 & 0.195 & 0.197 \\
 & 16 & 0.198 & 0.145 & 0.162 & 0.150 \\
 & 20 & 0.118 & 0.105 & 0.075 & 0.082 \\
 & 24 & 0.057 & 0.090 & 0.078 & 0.092 \\
\bottomrule
\end{tabular}

%% file: figures/rebuttal/tab_variants.tex
\begingroup\small\setlength{\tabcolsep}{4pt}
\begin{longtable}{llrccccrc}
\caption{Graph topology and serialization variants, Qwen2.5-7B, budget $16$. SBM: two communities; planted cycle: a cycle on $3$ to $n/2$ random nodes plus sparse noise (cycle detection is omitted since the answer is always yes); adjacency lists and letter labels follow \citet{herbst2025lost} (edge reordering acts trivially on adjacency lists, so only relabeling moves the prompt).}\label{tab:app-variants}\\
\toprule
Model & Task & $n$ & $N$ & Greedy & SC-16 & Sym.-16 & Gain & flip \\ \midrule \endfirsthead
\toprule
Model & Task & $n$ & $N$ & Greedy & SC-16 & Sym.-16 & Gain & flip \\ \midrule \endhead
\bottomrule \endfoot
SBM & edge count & 16 & 100 & 0.70 & 0.92 & 0.97 & $\mathbf{+4.8}$ $[+0.2,+10.2]$ & 0.78 \\
 & edge count & 24 & 100 & 0.40 & 0.77 & 0.88 & $\mathbf{+10.9}$ $[+3.5,+18.5]$ & 1.00 \\
 & edge existence & 16 & 100 & 0.90 & 0.97 & 1.00 & $\mathbf{+2.8}$ $[+0.3,+6.2]$ & 0.77 \\
 & edge existence & 24 & 100 & 0.86 & 0.97 & 0.99 & $+2.2$ $[-0.3,+5.5]$ & 0.85 \\
 & node degree & 16 & 100 & 0.50 & 0.48 & 0.45 & $-2.1$ $[-10.8,+6.2]$ & 0.91 \\
 & node degree & 24 & 100 & 0.20 & 0.19 & 0.15 & $-4.6$ $[-10.7,+0.8]$ & 0.97 \\
 & components & 16 & 100 & 0.57 & 0.74 & 0.79 & $\mathbf{+5.0}$ $[+1.7,+8.9]$ & 0.99 \\
 & components & 24 & 100 & 0.52 & 0.79 & 0.78 & $-0.1$ $[-2.9,+2.8]$ & 1.00 \\
 & connected & 16 & 100 & 0.77 & 0.79 & 0.82 & $+2.9$ $[-2.7,+8.9]$ & 0.79 \\
 & connected & 24 & 100 & 0.57 & 0.61 & 0.57 & $-3.8$ $[-8.2,+0.2]$ & 0.75 \\
 & cycle & 16 & 100 & 0.61 & 0.55 & 0.52 & $\mathbf{-2.8}$ $[-6.1,-0.2]$ & 0.48 \\
 & cycle & 24 & 100 & 0.56 & 0.50 & 0.50 & $+0.0$ $[+0.0,+0.0]$ & 0.38 \\
 & triangles & 16 & 100 & 0.13 & 0.14 & 0.11 & $-2.6$ $[-7.6,+1.8]$ & 0.99 \\
 & triangles & 24 & 100 & 0.01 & 0.02 & 0.03 & $+0.9$ $[-1.8,+2.9]$ & 1.00 \\
\midrule
planted cycle & edge count & 16 & 100 & 0.96 & 0.98 & 1.00 & $+2.0$ $[+0.0,+5.0]$ & 0.54 \\
 & edge count & 24 & 100 & 0.50 & 0.89 & 0.95 & $\mathbf{+6.8}$ $[+1.0,+13.1]$ & 0.84 \\
 & edge existence & 16 & 100 & 0.95 & 0.99 & 1.00 & $+0.5$ $[+0.0,+1.5]$ & 0.54 \\
 & edge existence & 24 & 100 & 0.90 & 0.98 & 1.00 & $+1.7$ $[+0.0,+4.3]$ & 0.61 \\
 & node degree & 16 & 100 & 0.82 & 0.80 & 0.83 & $+3.0$ $[-3.5,+9.8]$ & 0.68 \\
 & node degree & 24 & 100 & 0.56 & 0.56 & 0.55 & $-0.3$ $[-9.2,+8.4]$ & 0.84 \\
 & components & 16 & 100 & 0.21 & 0.23 & 0.27 & $+3.7$ $[-1.3,+8.8]$ & 1.00 \\
 & components & 24 & 100 & 0.16 & 0.31 & 0.30 & $-1.2$ $[-5.0,+1.9]$ & 1.00 \\
 & connected & 16 & 100 & 0.72 & 0.83 & 0.85 & $+2.0$ $[-2.5,+6.7]$ & 0.90 \\
 & connected & 24 & 100 & 0.67 & 0.66 & 0.68 & $+2.4$ $[-2.2,+6.7]$ & 0.89 \\
 & triangles & 16 & 100 & 0.13 & 0.18 & 0.24 & $+5.2$ $[-1.6,+11.7]$ & 1.00 \\
 & triangles & 24 & 100 & 0.07 & 0.11 & 0.12 & $+1.0$ $[-4.7,+5.9]$ & 1.00 \\
\midrule
adjacency lists & edge count & 16 & 100 & 0.15 & 0.13 & 0.09 & $-4.2$ $[-9.4,+0.6]$ & 0.99 \\
 & edge existence & 16 & 100 & 0.98 & 1.00 & 1.00 & $+0.0$ $[+0.0,+0.0]$ & 0.24 \\
 & node degree & 16 & 100 & 0.95 & 0.97 & 1.00 & $\mathbf{+3.3}$ $[+0.3,+7.2]$ & 0.45 \\
 & connected & 16 & 100 & 0.94 & 0.97 & 0.98 & $\mathbf{+1.4}$ $[+0.1,+3.2]$ & 0.36 \\
\midrule
letter labels & edge count & 16 & 100 & 0.95 & 0.98 & 1.00 & $+1.6$ $[-0.0,+4.2]$ & 0.43 \\
 & edge existence & 16 & 100 & 0.97 & 1.00 & 1.00 & $+0.0$ $[+0.0,+0.0]$ & 0.52 \\
 & node degree & 16 & 100 & 0.64 & 0.66 & 0.68 & $+2.5$ $[-6.0,+10.7]$ & 0.78 \\
 & connected & 16 & 100 & 0.71 & 0.79 & 0.83 & $+4.2$ $[-0.8,+9.3]$ & 0.93 \\
\end{longtable}
\endgroup

%% file: figures/rebuttal/tab_premises.tex
\begingroup\small\setlength{\tabcolsep}{4pt}
\begin{longtable}{llrccccrc}
\caption{Deduction from $k-1$ adjacent height comparisons among $k$ people, Qwen2.5-7B, budget $16$, under premise reordering ($S_{k-1}$), statement flipping ($\mathbb Z_2^{k-1}$), and their product. Each group is compared against its own baselines, since the starting prompt is a random element of that orbit.}\label{tab:app-premises}\\
\toprule
Model & Task & $k$ & $N$ & Greedy & SC-16 & Sym.-16 & Gain & flip \\ \midrule \endfirsthead
\toprule
Model & Task & $k$ & $N$ & Greedy & SC-16 & Sym.-16 & Gain & flip \\ \midrule \endhead
\bottomrule \endfoot
reorder & tallest & 5 & 100 & 0.98 & 1.00 & 1.00 & $+0.0$ $[+0.0,+0.0]$ & 0.09 \\
 & tallest & 8 & 100 & 0.93 & 0.99 & 1.00 & $+0.5$ $[+0.0,+1.5]$ & 0.72 \\
 & tallest & 11 & 100 & 0.83 & 0.94 & 1.00 & $\mathbf{+5.8}$ $[+1.8,+10.8]$ & 0.89 \\
 & $j$-th tallest & 5 & 100 & 0.90 & 0.99 & 1.00 & $+0.9$ $[-0.2,+2.9]$ & 0.70 \\
 & $j$-th tallest & 8 & 100 & 0.63 & 0.84 & 0.92 & $\mathbf{+7.5}$ $[+0.5,+15.1]$ & 0.98 \\
 & $j$-th tallest & 11 & 100 & 0.49 & 0.64 & 0.72 & $+7.8$ $[-0.7,+15.9]$ & 1.00 \\
 & comparison & 5 & 100 & 0.93 & 0.99 & 1.00 & $+1.3$ $[-0.2,+3.8]$ & 0.38 \\
 & comparison & 8 & 100 & 0.85 & 0.94 & 0.97 & $+2.8$ $[-1.1,+7.3]$ & 0.70 \\
 & comparison & 11 & 100 & 0.80 & 0.88 & 0.90 & $+2.6$ $[-1.5,+6.9]$ & 0.75 \\
\midrule
flip & tallest & 5 & 100 & 0.95 & 1.00 & 1.00 & $+0.0$ $[+0.0,+0.0]$ & 0.29 \\
 & tallest & 8 & 100 & 0.74 & 0.95 & 0.99 & $\mathbf{+4.0}$ $[+0.4,+8.1]$ & 0.90 \\
 & tallest & 11 & 100 & 0.61 & 0.77 & 0.86 & $\mathbf{+9.6}$ $[+2.6,+16.8]$ & 0.99 \\
 & $j$-th tallest & 5 & 100 & 0.80 & 0.93 & 0.99 & $\mathbf{+6.5}$ $[+2.1,+11.7]$ & 0.91 \\
 & $j$-th tallest & 8 & 100 & 0.32 & 0.44 & 0.60 & $\mathbf{+16.4}$ $[+8.0,+24.7]$ & 1.00 \\
 & $j$-th tallest & 11 & 100 & 0.18 & 0.28 & 0.30 & $+2.8$ $[-4.8,+10.1]$ & 1.00 \\
 & comparison & 5 & 100 & 0.98 & 1.00 & 1.00 & $-0.2$ $[-0.4,+0.0]$ & 0.29 \\
 & comparison & 8 & 100 & 0.72 & 0.80 & 0.88 & $\mathbf{+8.3}$ $[+2.1,+15.0]$ & 0.81 \\
 & comparison & 11 & 100 & 0.71 & 0.74 & 0.81 & $+6.6$ $[-0.2,+13.8]$ & 0.88 \\
\midrule
both & tallest & 5 & 100 & 1.00 & 1.00 & 1.00 & $+0.0$ $[+0.0,+0.0]$ & 0.51 \\
 & tallest & 8 & 100 & 0.75 & 0.86 & 1.00 & $\mathbf{+13.9}$ $[+7.0,+20.9]$ & 0.98 \\
 & tallest & 11 & 100 & 0.51 & 0.79 & 0.99 & $\mathbf{+19.5}$ $[+11.9,+27.3]$ & 1.00 \\
 & $j$-th tallest & 5 & 100 & 0.81 & 0.97 & 1.00 & $\mathbf{+3.0}$ $[+0.4,+6.5]$ & 0.92 \\
 & $j$-th tallest & 8 & 100 & 0.35 & 0.50 & 0.64 & $\mathbf{+14.1}$ $[+5.6,+22.8]$ & 1.00 \\
 & $j$-th tallest & 11 & 100 & 0.21 & 0.33 & 0.32 & $-0.9$ $[-9.1,+6.8]$ & 1.00 \\
 & comparison & 5 & 100 & 0.95 & 0.98 & 1.00 & $+1.9$ $[-0.1,+4.8]$ & 0.46 \\
 & comparison & 8 & 100 & 0.82 & 0.88 & 0.95 & $\mathbf{+6.8}$ $[+1.6,+12.1]$ & 0.88 \\
 & comparison & 11 & 100 & 0.70 & 0.75 & 0.83 & $\mathbf{+8.5}$ $[+1.6,+15.2]$ & 0.93 \\
\end{longtable}
\endgroup

%% file: figures/rebuttal/tab_icl.tex
\begingroup\small\setlength{\tabcolsep}{4pt}
\begin{longtable}{llrccccrc}
\caption{Few-shot sentiment classification (SST-2) with $k$ demonstrations, rewritten by permuting the demonstrations ($S_k$), Qwen2.5-7B, budget $16$.}\label{tab:app-icl}\\
\toprule
Model & Task & $k$ & $N$ & Greedy & SC-16 & Sym.-16 & Gain & flip \\ \midrule \endfirsthead
\toprule
Model & Task & $k$ & $N$ & Greedy & SC-16 & Sym.-16 & Gain & flip \\ \midrule \endhead
\bottomrule \endfoot
SST-2 & SST-2 & 4 & 200 & 0.95 & 0.96 & 0.96 & $-0.3$ $[-1.1,+0.3]$ & 0.07 \\
 & SST-2 & 8 & 200 & 0.96 & 0.97 & 0.97 & $+0.3$ $[-0.3,+1.2]$ & 0.08 \\
 & SST-2 & 16 & 200 & 0.95 & 0.97 & 0.97 & $+0.2$ $[+0.0,+0.7]$ & 0.06 \\
\end{longtable}
\endgroup

%% file: figures/rebuttal/tab_paraphrase.tex
\begingroup\small\setlength{\tabcolsep}{4pt}
\begin{longtable}{llrccccrc}
\caption{Rewritings sampled from a paraphrasing model instead of the group, Qwen2.5-7B answering, budget $16$; rows are labeled by the paraphraser (cross-family: Phi-4-mini; same-model: Qwen2.5-7B). Instances with fewer than one surviving paraphrase are dropped, hence $N<100$ at $n=24$. Baselines (Greedy, SC-16) are computed on a paraphrase, so they differ from those of the exact group on the same instances (\cref{tab:app-main-full}).}\label{tab:app-paraphrase}\\
\toprule
Model & Task & $n$ & $N$ & Greedy & SC-16 & Sym.-16 & Gain & flip \\ \midrule \endfirsthead
\toprule
Model & Task & $n$ & $N$ & Greedy & SC-16 & Sym.-16 & Gain & flip \\ \midrule \endhead
\bottomrule \endfoot
cross-family & edge count & 12 & 100 & 0.85 & 0.89 & 0.99 & $\mathbf{+9.8}$ $[+3.8,+15.9]$ & 0.80 \\
 & edge count & 24 & 78 & 0.46 & 0.68 & 0.74 & $+6.5$ $[-3.0,+16.4]$ & 0.96 \\
 & edge existence & 12 & 100 & 0.96 & 0.96 & 0.98 & $+1.2$ $[-0.4,+3.7]$ & 0.30 \\
 & edge existence & 24 & 72 & 0.94 & 0.97 & 0.98 & $+0.8$ $[-0.6,+2.8]$ & 0.38 \\
 & node degree & 12 & 100 & 0.65 & 0.66 & 0.62 & $-4.5$ $[-9.6,+0.2]$ & 0.55 \\
 & node degree & 24 & 77 & 0.52 & 0.53 & 0.55 & $+1.8$ $[-3.6,+7.5]$ & 0.60 \\
 & components & 12 & 100 & 0.47 & 0.59 & 0.62 & $+2.6$ $[-4.1,+9.7]$ & 0.93 \\
 & components & 24 & 73 & 0.49 & 0.53 & 0.55 & $+2.0$ $[-1.6,+5.7]$ & 0.75 \\
 & connected & 12 & 100 & 0.75 & 0.80 & 0.78 & $-2.1$ $[-6.9,+2.2]$ & 0.74 \\
 & connected & 24 & 86 & 0.62 & 0.55 & 0.57 & $+2.1$ $[-2.9,+7.3]$ & 0.58 \\
 & cycle & 12 & 100 & 0.62 & 0.58 & 0.59 & $+0.6$ $[-1.2,+2.2]$ & 0.39 \\
 & cycle & 24 & 84 & 0.46 & 0.41 & 0.41 & $-0.3$ $[-1.4,+0.5]$ & 0.44 \\
 & triangles & 12 & 100 & 0.20 & 0.27 & 0.21 & $-5.9$ $[-12.7,+0.9]$ & 0.99 \\
 & triangles & 24 & 75 & 0.11 & 0.08 & 0.10 & $+1.3$ $[-3.5,+5.8]$ & 0.97 \\
\midrule
same-model & edge count & 24 & 60 & 0.47 & 0.84 & 0.85 & $+0.7$ $[-6.4,+7.8]$ & 0.80 \\
 & edge existence & 24 & 58 & 0.91 & 0.98 & 0.98 & $-0.1$ $[-3.9,+4.4]$ & 0.28 \\
 & node degree & 24 & 61 & 0.64 & 0.60 & 0.65 & $\mathbf{+4.6}$ $[+0.4,+9.6]$ & 0.34 \\
\end{longtable}
\endgroup

%% file: sections/notation.tex
\section{Notation}
\label{app:notation}

This table collects the notation used in the paper and the place where each symbol is introduced.

{\small
\begin{longtable}{@{}>{\raggedright\arraybackslash}p{0.27\linewidth}>{\raggedright\arraybackslash}p{0.52\linewidth}>{\raggedright\arraybackslash}p{0.14\linewidth}@{}}
    \toprule
    Symbol & Meaning & Introduced \\
    \midrule
    \endfirsthead
    \toprule
    Symbol & Meaning & Introduced \\
    \midrule
    \endhead
    \bottomrule
    \endlastfoot
    $\mathcal X$ & space of prompts & \cref{sec:formulation} \\
    $\mathcal C$, $C=|\mathcal C|$ & set of answer classes and its size & \cref{sec:formulation} \\
    $y^\star:\mathcal X\to\mathcal C$ & target function, invariant: $y^\star(g\cdot x)=y^\star(x)$ & \cref{sec:formulation} \\
    $G$, $|G|$, $e$ & finite group of rewritings, its order, and its identity & \cref{sec:formulation} \\
    $g\cdot x$ & rewriting (serialization) of prompt $x$ by $g\in G$ & \cref{sec:formulation} \\
    $G\cdot x$ & orbit of $x$: all rewritings of the same instance & \cref{sec:formulation} \\
    $h$ & starting rewriting from which inference is invoked & \cref{eq:sparse-operator} \\
    $\rho(g)$ & permutation of answer classes induced by $g$ (equivariant targets) & \cref{app:preliminaries} \\
    $X$ & prompt drawn from the task distribution & \cref{sec:decision-corollary} \\
    \midrule
    $p_{\theta,\tau}(\cdot\mid x)$ & frozen model: distribution of reasoning traces given prompt $x$ & \cref{eq:score-field} \\
    $\tau$ & generation temperature & \cref{eq:score-field} \\
    $\omega\in\Omega$ & a sampled reasoning trace and the trace space & \cref{eq:score-field} \\
    $V_x(g,\omega)\in\mathcal Z$ & response field: response of trace $\omega$ at rewriting $g\cdot x$ & \cref{eq:score-field} \\
    $a_x(g,\omega)$, $r_x(g,\omega)$ & answer extracted from a trace and its verifier score & \cref{eq:decoder-verifier} \\
    $\mathcal D$, $\mathcal H$ & decoding rule and the real Hilbert space of its outputs & \cref{eq:decoded-orbit-field} \\
    $F_x^{\mathcal D}(g)$ & decoded orbit field: output of $\mathcal D$ at rewriting $g$ & \cref{eq:decoded-orbit-field} \\
    $F_x^{\rm SC}$, $F_x^{\rm logit}$, $F_x^{\beta_D}$ & self consistency, expected logits, verifier weighted decoding & \cref{sec:formulation} \\
    $\beta_D$ & decoder (verifier) inverse temperature & \cref{eq:decoder-verifier} \\
    $q_x^{\mathcal D}=\mathsf P_GF_x^{\mathcal D}$ & fully symmetrized response & \cref{eq:full-symmetrization} \\
    $\widehat F_{x,\xi}$, $\xi$ & field of a realized randomized decoder and its seed & \cref{app:preliminaries} \\
    $n_D$ & number of traces per rewriting in finite self consistency & \cref{app:joint-sampling} \\
    \midrule
    $L^2(G;\mathcal H)$ & $\mathcal H$-valued functions on $G$ with the uniform inner product & \cref{app:preliminaries} \\
    $S=\{\!\{s_1,\ldots,s_{|S|}\}\!\}$ & multiset of rewritings, drawn i.i.d.\ from $\Unif(G)$ & \cref{eq:sparse-operator} \\
    $\mathsf A_S$ & sparse rewriting operator $(\mathsf A_Sf)(h)=\frac1{|S|}\sum_i f(s_ih)$ & \cref{eq:sparse-operator} \\
    $\mathsf P_G$ & full group average, the orthogonal projection onto constants & \cref{eq:reynolds-projector} \\
    $\mathsf T_s$ & translation operator $(\mathsf T_sf)(h)=f(sh)$ & \cref{eq:translation-operator} \\
    $\|\cdot\|_{\mathrm{op}}$ & operator norm on $L^2(G)$ & \cref{eq:l2-operator-norm} \\
    $\delta$ & failure probability of the random draw of $S$ & \cref{thm:sparse-reynolds} \\
    $\pi$, $\mathcal U(\mathcal H)$ & unitary representation of $G$ and the unitary group of $\mathcal H$ & \cref{app:preliminaries} \\
    $\mathcal H^G$, $\mathsf P_\pi$ & invariant vectors of $\pi$ and the projection onto them & \cref{app:preliminaries} \\
    $\lambda$ & left regular representation of $G$ & \cref{app:preliminaries} \\
    $\widehat G$, $d_\rho$ & irreducible unitary representations and the dimension of $\rho\in\widehat G$ & \cref{app:preliminaries} \\
    \midrule
    $c_G(X)$ & decision of the fully symmetrized model, $\arg\max_c(q_X)_c$ & \cref{sec:decision-corollary} \\
    $\widehat c_S(h\cdot X)$ & decision of sparse symmetrization from starting rewriting $h$ & \cref{sec:decision-corollary} \\
    $\Delta_X$ & margin between the two largest coordinates of $q_X$ & \cref{sec:decision-corollary} \\
    $R_G$, $R_S$ & error of full and of sparse symmetrization, $h\sim\Unif(G)$ & \cref{sec:decision-corollary} \\
    $t$ & margin threshold in the error decomposition & \cref{cor:margin} \\
    \midrule
    $n$ & number of nodes of a graph; node relabelings form $G=S_n$, $|G|=n!$ & \cref{sec:experiments} \\
    $\mathbb Z_2^d$ & group of $d$ binary flips, $|G|=2^d$ (the family of the lower bound) & \cref{sec:experiments} \\
    $\eta_S(F)$ & relative error $\|\mathsf A_SF-\mathsf P_GF\|/\|F-\mathsf P_GF\|$ & \cref{sec:experiments} \\
    $k$ & budget of forward passes per question & \cref{sec:experiments} \\
    $m$, $r$ & number of rewritings and number of sampled answers per rewriting, $k=m\cdot r$ & \cref{tab:app-allocation} \\
\end{longtable}
}